\documentclass[sigconf]{acmart}

\AtBeginDocument{%
  }

\copyrightyear{2025}
\acmYear{2025}
\setcopyright{cc}
\setcctype{by}
\acmConference[SIGSPATIAL '25]{The 33rd ACM International Conference on Advances in Geographic Information Systems}{November 3--6, 2025}{Minneapolis, MN, USA}
\acmBooktitle{The 33rd ACM International Conference on Advances in Geographic Information Systems (SIGSPATIAL '25), November 3--6, 2025, Minneapolis, MN, USA}
\acmDOI{10.1145/3748636.3764172}
\acmISBN{979-8-4007-2086-4/2025/11}

\usepackage{enumitem}
\usepackage{soul}
\usepackage{multirow}
\usepackage{tikz}
\usetikzlibrary{matrix,arrows.meta}
\usepackage{graphicx} 
\usepackage{subcaption}
\usepackage{float} 
\usepackage{hyperref}
\usepackage[endLComment=]{algpseudocodex}
\usepackage[ruled]{algorithm}
\usepackage{caption}
\usepackage{etoolbox}
\AtBeginEnvironment{align}{\setlength{\abovedisplayskip}{4pt} \setlength{\belowdisplayskip}{4pt}}
\usepackage{titlesec}
\titlespacing*{\section}{0pt}{5pt}{1pt}
\usepackage{titlesec}
\titlespacing*{\subsection}{0pt}{5pt}{0pt}
\usepackage{titlesec}
\titlespacing*{\subsubsection}{0pt}{3pt}{2pt}

\begin{document}

\title{HydroGAT: Distributed Heterogeneous Graph Attention Transformer for Spatiotemporal Flood Prediction}

\author{Aishwarya Sarkar$^*$, Autrin Hakimi$^*$, Xiaoqiong Chen$^*$, Hai Huang$^*$, Chaoqun Lu$^*$, Ibrahim Demir$^{\dagger}$, Ali Jannesari$^*$}

\affiliation{
  \institution{$^*$Iowa State University, Ames, IA \{asarkar1, ahakimi, xqchen, haihuang, clu, jannesari\}@iastate.edu}
  \city{}
  \state{}
  \country{}
}
  \affiliation{
  \institution{$^{\dagger}$Tulane University, New Orleans, LA, idemir@tulane.edu}
  \city{}
  \state{}
  \country{}
}

\renewcommand{\shortauthors}{Sarkar et al.}
\newcommand{\note}[1]{\textcolor{red}{{[}#1{]}}}

\begin{abstract}
Accurate flood forecasting remains a critical challenge for water‐resource management, as it demands simultaneous modeling of local, time-varying runoff drivers (e.g., rainfall-induced peaks, baseflow trends) and complex spatial interactions across a river network. Traditional data-driven approaches, such as convolutional networks and sequence-based models, ignore topological information about the region. Graph Neural Networks (GNNs), in contrast, propagate information exactly along the river network, making them ideal for learning hydrological routing. However, state-of-the-art GNN-based flood prediction models still collapse pixels to coarse catchment polygons because the cost of training explodes with graph size and higher resolution. Furthermore, most existing methods treat spatial and temporal dependencies separately, either applying GNNs solely on spatial graphs or transformers purely on temporal sequences, thus failing to simultaneously capture spatiotemporal interactions critical for accurate flood prediction. 
To address these limitations, we introduce a heterogenous basin graph to represent every land and river pixel as a node connected by both physical hydrological flow directions as well as inter-catchment relationships. We also propose HydroGAT, a novel spatiotemporal network that adaptively learns both local temporal importance as well as most influential upstream locations. Evaluated in two Midwestern US basins and across five baseline architectures, our model achieves higher NSE (up to 0.97), improved KGE (up to 0.96), and low bias (PBIAS within $\pm$ 5\%) in hourly discharge prediction, while offering interpretable attention maps that reveal sparse, structured intercatchment influences. To support high-resolution basin-scale training, we develop a distributed data-parallel pipeline that scales efficiently up to 64 NVIDIA A100 GPUs on NERSC Perlmutter supercomputer, demonstrating up to 15$\times$ speedup across machines. Our code is available at \url{https://github.com/swapp-lab/HydroGAT}.
\end{abstract}

\vspace{-3em}

\vspace{-1em}

\maketitle

\section{Introduction}\label{sec:intro}
Accurate river discharge forecasting is critical for mitigating flood risks, safeguarding infrastructure, and supporting ecological sustainability ~\cite{yildirim2022flood}. This task is especially crucial in regions vulnerable to extreme rainfall or snowmelt, where timely predictions can enable rapid interventions to protect both infrastructure and human lives~\cite{alabbad2023web}. The Midwest of the United States, particularly Iowa, exemplifies such vulnerable regions. As one of the world's three major black soil regions, Iowa serves as a key agricultural hub, yet it frequently experiences flooding driven by a combination of extreme weather events and intensive agricultural practices. Unique land-use features, such as high-density tile drainage networks~\cite{valayamkunnath2020mapping} and the prairie pothole landscape~\cite{bishop1998iowa}, strongly influence local hydrological responses. Moreover, seasonal crop phenology modulates evapotranspiration, while extended winters introduce complex rainfall-snowmelt interactions, resulting in highly dynamic and heterogeneous flood behaviors in the region. However, accurate discharge forecasting remains highly challenging due to the complex and nonlinear interactions among meteorological, hydrological, and topographic factors. Traditional physics-based models provide physical interpretability but require extensive calibration and struggles on ungauged basins~\cite{article}. In contrast, data-driven models, especially machine learning (ML) approaches, offer flexibility by directly learning from historical data~\cite{sit2020comprehensive} and often achieve higher predictive accuracy. Nevertheless, many ML models treat hydrological systems as purely statistical processes, frequently overlooking the underlying topology of catchments, such as drainage networks and hillslope connectivity. This limitation affects their ability to generalize under non-stationary conditions (e.g., climate change, land-use shifts) and restricts their interpretability~\cite{kumar2024data}.

\begin{figure}[h] 
  \centering
  \vspace{-1em}
  \includegraphics[width=0.9\columnwidth]{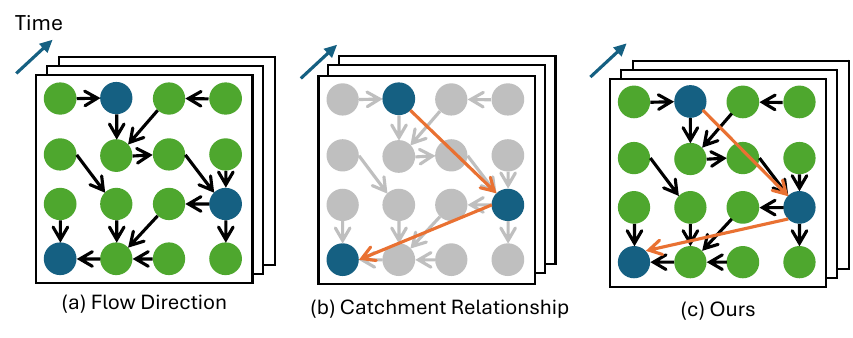}
  \caption{Heterogeneous graph construction. (a) Flow-direction edges (black) capture direct hydrological connectivity; (b) catchment relationship edges (orange) represent indirect dependencies across distant upstream-downstream catchments; (c) our dual-edge graph combines both edge types over time, modeling interactions between river station (blue) and land (green) nodes.}\vspace{-1em}
  \label{fig:dual_edge}
\end{figure}

In flood forecasting, explicitly incorporating both channel networks and hillslope processes is critical, as floods are inherently shaped by the spatial organization of the landscape. Graph Neural Networks (GNNs), which can naturally represent spatially distributed hydrological systems as graphs, offer a promising framework to capture these complex interactions by modeling both river channels and hillslope connectivity. This enables finer-scale predictions at the pixel level while preserving the underlying topological structure of the basin, thereby enhancing both accuracy and interpretability in flood forecasting. Existing graph-based models represent basins as graphs using three approaches \cite{willard2025time}: (i) \textit{similarity-based} edges, which link sites based on attributes or proximity (site characteristics \cite{sun2021explore}, spatial proximity \cite{sun2021explore}, or both \cite{xiang2021high}), (ii) \textit{physics-based} edges, which encode real water routing paths like upstream-downstream links \cite{jia2021physics, sit2021shorttermhourlystreamflowprediction}, and (iii) \textit{adaptive} edges, where the graph structure is learned during training \cite{sun2022graph}. Similarity edges are attractive for large-scale studies because they can link basins that are far apart and thus bypass the need for explicit routing, but they add no physical knowledge beyond what could be captured by simply concatenating attributes. Physics-based edges encode true hydrological routing across the landscape but have been mostly limited to stream networks. As a result, most prior works only include the wet channels, i.e., each node in the graph is either a gauge station, a reach centroid, or a catchment polygon outlet, while critical land processes like snow accumulation, soil-moisture storage, and infiltration excess are excluded. Consequently, the network is forced to infer runoff generation signals indirectly from the sparse channel nodes that overlook fine-grained heterogeneity in hydrological processes. To overcome this limitation, we redefine the graph to promote every pixel to a node, treating rivers and surrounding land surface within a single, unified graph following the idea of \cite{xiang2021high}, and introduce a heterogeneous topology with physically distinct edge types (Fig.~\ref{fig:dual_edge}). We further introduce an attention-based spatiotemporal network that reveals which upstream pixels and past timesteps drive each prediction, directly tackling the interpretability challenge that plagues deep learning models, an issue made worse when spatial processes are implicit rather than explicit \cite{articleJiang}. 

Moving to a high-resolution basin graph increases both node count per basin and input sequence length by 10-100$\times$, inflating memory and compute demands beyond a single GPU. This often forces coarse graph approximations or pruning the graph aggressively, which sacrifices spatial information in most operational flood prediction systems. To preserve full resolution and keep training manageable, we exploit modern high-performance computing (HPC) systems to distribute the data across machines, allowing each GPU to process a distinct temporal slice of the basin.
This approach scales almost linearly across machines, reducing multi-day single-GPU jobs to a few hours while preserving full spatial resolution. In summary, our contributions are:
\begin{enumerate}
    \item Heterogenous graph representation of basins where each pixel (river or land) is a node, with two edge types (flow-direction and inter-catchment relationship) to preserve hydrological routing and landscape connectivity.
    \item HydroGAT, an interpretable spatiotemporal attention architecture that combines a transformer-based temporal encoder with dual GAT-based spatial branches, fused by a gated learnable parameter to capture when and where to focus.
    \item A distributed memory (multi-GPU multi-machine) pipeline to accelerate training on high-resolution basin graphs.
    \item Evaluation on two Midwest U.S. basins against five state-of-the-art baselines, with ablation studies to assess component contributions and the model's reasoning.
\end{enumerate}
\section{Related Works}\label{sec:related} 
Classical statistical models such as ARIMA and multiple linear regression~\cite{wong2020flood,fashae2019comparing,otchere2021application} have been used to model streamflow, but they are limited in capturing nonlinear dynamics. Deep learning models, especially recurrent neural networks (RNNs) like Long Short-Term Memory (LSTM) and Gated Recurrent Units (GRUs), have shown strong predictive performance~\cite{sit2020comprehensive} by learning complex temporal dependencies from data; however, they still face challenges in modeling long-range dependencies and incorporating spatial structure. To address this, graph neural networks (GNNs) have recently been adopted to embed river basin topology directly into data-driven models. Applications range from drainage pattern recognition~\cite{yu2022recognition}, groundwater level prediction~\cite{bai2023graph}, rainfall-runoff or streamflow prediction~\cite{feng2022graph, kratzert2021large, sit2021shorttermhourlystreamflowprediction, sun2021explore, zhao2020joint, sit2023spatial}, to stream and lake temperature prediction~\cite{stalder2021probabilistic}. For example,~\cite{feng2022graph} proposes a spatiotemporal network on a graph constructed solely from gauge stations that combines Chebyshev GCNs~\cite{defferrard2016convolutional} and LSTMs with a single head attention mechanism. This is far less expressive than modern multi-head self-attention as it cannot capture multiple, competing spatiotemporal dependencies. Similarly, \cite{kratzert2021large} runs separate LSTMs per sub-basin to generate local estimates and then passes these node outputs through a simple edge-wise routing network, so spatial interactions are learned only post-hoc, where the model cannot simultaneously learn coupled runoff-routing dynamics. Other works like \cite{sit2021shorttermhourlystreamflowprediction} propose a convolutional GRU network that uses upstream gauge connectivity for streamflow prediction. \cite{sun2021explore} proposes a GraphWaveNet architecture (originally proposed in \cite{wu2019graph}) with diffusion convolutions and dilated temporal CNNs to a dynamically built $k$-nearest-neighbor catchment graph. \cite{zhao2020joint} proposes three parallel graphs using hydraulic distance, Euclidean distance, and discharge correlation, and combines them with manually chosen weights. However, their nodes are still limited to gauge stations, and the static fusion coefficients prevent the model from learning how much each relation should matter under different flow regimes. \cite{xiang2022fully} proposes a graph-informed rainfall-runoff model that uses hydrological flow direction to define parameterized $k$-hop neighborhoods for aggregation. \cite{zanfei2022graph} proposed GCRNN (originally proposed in \cite{ruiz2019gatedgraphconvolutionalrecurrent}) for streamflow prediction. 

While these works demonstrate the growing potential of GNNs in flood prediction, most rely on sparse spatial inputs (e.g., gauge stations only) and overlook the influence of land surface processes. In agriculture-dominated watersheds, water management practices on land (such as tile drainage) and crop growth-driven seasonal patterns of water balance play a central role in shaping water flow dynamics. Lack of land surface information in the graph structure in most existing models limits their ability to capture the full spatial heterogeneity of runoff generation. Our work addresses this gap by incorporating both river and land nodes into the GNN framework, allowing the model to learn interactions across the coupled land-river system.

\begin{figure*}[!t]
  \centering
  \includegraphics[width=\textwidth]{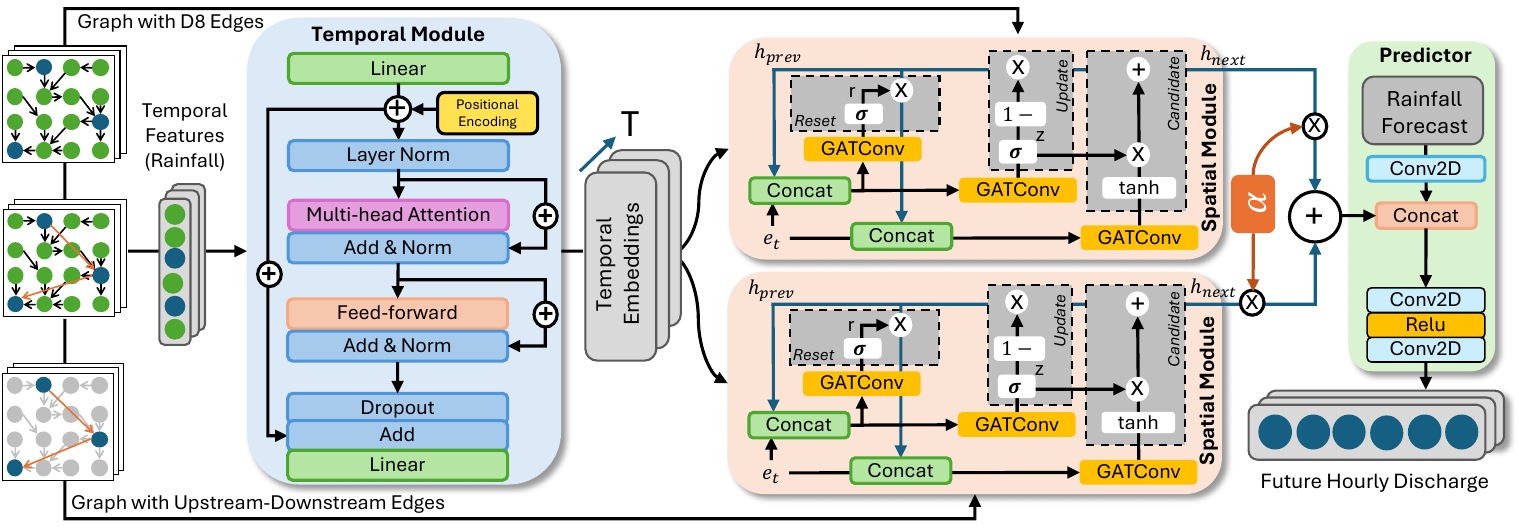}
  \caption{Architecture of HydroGAT. A transformer-based temporal module (blue) combined with two spatial modules (orange) based on GAT-based GRUs over flow and catchment edges. Temporal embeddings are combined with spatiotemporal hidden states and rainfall forecasts for downstream prediction via a convolutional predictor (green). Note: the updated hidden state $h_{next}$ is used as $h_{prev}$ recursively. After processing the full temporal window, the final hidden state is passed to the Predictor.}
  \label{fig:architecture}
\end{figure*}
\section{Overview of HydroGAT}\label{sec:methods}
We formally define the problem in \S\ref{subsec:prelim}, introduce our model HydroGAT in \S\ref{subsec:temp}-\S\ref{subsec:spatial}, and discuss our distributed pipeline in \S\ref{subsec:dist}.
\subsection{Preliminaries}\label{subsec:prelim}
We represent a river basin as a graph \(\mathcal{G}=(\mathcal{V},\mathcal{E},\mathbf{X})\), where nodes \(\mathcal{V}\) are uniform raster cells (grids/pixels of 4 km$^2$ extracted from gridded precipitation), each with its own time series and edges \(\mathcal{E}\) (flow-direction and catchment edges) encode hydrological connections, forming the heterogeneous graph in Fig.~\ref{fig:dual_edge}.
Then, \(\mathbf{X^{t}}=\bigl[\mathbf x^{\,t-T+1},\,\mathbf x^{\,t-T+2},\,\dots,\,\mathbf x^{\,t}\bigr] \in \mathbb{R}^{|\mathcal{V}| \times T \times F}\) encodes historical precipitation and discharge features ($F$) over a time window of length \(T\). In flood prediction modeling, the task is to learn a function \(f\) that predicts future discharge at a subset of target nodes (gauge stations within the region) \(\mathcal{V}_\rho \subseteq \mathcal{V}\). Formally,
\begin{align}
  f(\mathbf X^{t}, \mathcal G;\theta)\;:\;(\mathbf X^{t},\mathcal G)\longrightarrow 
  \hat{\mathbf Y}^{\,t+\Delta t}_{\mathcal V_\rho}\in\mathbb R^{|\mathcal V_\rho|\times\Delta t}
\end{align}
where \(\hat{\mathbf{Y}}^{t+\Delta t}_{\mathcal{V}_\rho} \in \mathbb{R}^{|\mathcal{V}_\rho| \times \Delta t}\) represents the predicted discharge at target nodes over the next \(\Delta t\) timesteps, and \(\theta\) denotes the learnable model parameters. Then, for each target node, the prediction is given by:
\begin{align}
\hat{\mathbf{y}}_v^{\,t+\Delta t} = f_v\bigl(\mathbf X^{t},\,\mathcal G;\,\theta\bigr) \quad \forall v \in \mathcal{V}_\rho
\end{align}
\subsubsection{\textbf{Nodes}}\label{subsubsec:nodes-definition}
Unlike prior works limited to gauge stations or catchment outlets, our graph includes all land and river pixels as nodes for finer spatial granularity. Each node ($v \in V$) is assigned a multivariate time series $\mathbf{X}_v \in \mathbb{R}^{T \times F}$, where \(T\) is the input sequence length and \(F\) the number of input channels (e.g., precipitation, discharge if available). Precipitation inputs are available at all nodes, while both historical discharge (as inputs) and future discharge (as labels) are provided only at target nodes, \(\mathcal{V}_{\rho} \subseteq \mathcal{V}\). Nodes without discharge observations remain unlabeled and contribute solely to spatiotemporal feature propagation within the graph. Details of the study region are discussed in \S\ref{subsubsec:studyregion}.
\subsubsection{\textbf{Edges}}
To capture both physical transport and broader hydrological dependencies, we represent basins as heterogeneous graphs with two types of \textit{directed} edges.
\begin{enumerate}
    \item \textbf{Flow edges (\(\mathcal{E}_F\))} encode topographic water routing via the commonly used D8 algorithm \cite{yang2023study}. Each node \(u\) has a single outgoing edge to the neighbor \(v\) of steepest descent. Formally, $\mathcal{E}_F = \{ (u \rightarrow v) \mid v = \text{D8}(u)\}$.
    \item \textbf{Catchment edges (\(\mathcal{E}_C\))} encode physically routed upstream-downstream dependencies between target nodes along the river network. An edge \((u, v)\) in \(\mathcal{E}_C\) implies that a target node \(u\) may influence the discharge at target node \(v\), even in the absence of close proximity. This allows the model to learn shared responses to regional climatic inputs at basin-scale.
    \item \textbf{Self-loop edges} \((v,v)\) for every node \(v\in\mathcal V\)incorporate its own previous state into their update to capture local temporal dynamics. This is a common practice in message-passing models \cite{kipf2016semi}.
\end{enumerate}
Edge weights represent the context-specific importance of each connection. In our implementation, edge weights are learned implicitly via attention mechanisms of our model, described in \S\ref{subsec:spatial}.
\vspace{-1em}
\begin{algorithm}
{\small
\caption{\textsc{\textbf{Distributed Training of HydroGAT}} \\
\textbf{Input:} $\mathcal{G} (\mathcal{V}, \mathcal{E_{\text{F}} |\mathcal{E_\text{C}}})$: Heterogeneous basin graph, $\mathcal{V}_\rho \in \mathcal{V}$ : Set of target nodes (gauge stations), $\mathcal{D}(X,P)$: Training dataset, $X$: Past inputs (precipitation and discharge), $P$: forecasted rainfall, $\alpha$: Learnable weight, $T$: Input time window, $\Delta t$: Output time window \\
\textbf{Output:} ${\Psi}^*$: Trained model
}
\label{alg:model}
\begin{algorithmic}[1]
    \State \(\Psi\),\,\(\alpha\gets\) initialize all model parameters 
    \State $\mathcal{G}_{\text{flow}}, \mathcal{G}_{\text{catch}} \gets \mathcal{G}$ \Comment{Extract subgraphs}
    \For{$e$ in \textit{range(\texttt{\#epochs})} \textbf{in parallel}} \label{alg:begin_training} \Comment{Parallel training per trainer}
    \LComment{Each trainer samples a unique batch of training data on $\mathcal{G}$}
        \State $X, P$ $\gets$ \Call{Sequential\_Sampler}{$\mathcal{D}$, batch\_size, trainer\_id} 
        \State $E_{\text{seq}} \gets \mathrm{TemporalEncoder}(X)$\label{alg:temporal-encoder}  \Comment{Learn temporal embeddings}
        \State ${h}_v^{t-1}\gets h^{(0)}$  \Comment{Initialize hidden states; ${h}_v^{t-1}=h_{prev}$}
        \For{$t = 1$ to $T$} \Comment{One GRU-GAT update per time step} 
            \State \(e_t \gets E_{\text{seq}}^{t}\) \Comment{Temporal embedding at time \(t\)}
            \State \(h_\mathcal{V}^{(t,\text{flow})} \gets \mathrm{GRUGAT}\bigl(e_t,\;{h}_v^{t-1},\;\mathcal{G}_{\text{flow}}\bigr)\) \Comment{Update with $\mathcal{E_{\text{F}}}$ edges} \label{alg:gatflow}
            \State \(h_\mathcal{V_{\rho}}^{(t,\text{catch})} \gets \mathrm{GRUGAT}\bigl(e_t,\;{h}_v^{t-1},\;\mathcal{G}_{\text{catch}}\bigr)\) \Comment{Update with $\mathcal{E_{\text{C}}}$ edges} \label{alg:gatcatch}
            \State \(\alpha\gets\sigma(\alpha)\)  \Comment{Learnable fusion parameter} \label{alg:alphainit}
            \For{$v \in \mathcal{V}$} \Comment{Merge branches at each target} \label{alg:merge-start}
                \If{$v$ is a target node ($v\in\mathcal{V_{\rho}}$)}
                \State \(\tilde h^{(t)}_v \gets \alpha\odot h_{v}^{(t,\text{flow})} \;+\;(1-\alpha)\odot h_{v}^{(t,\text{catch})}\)
                \Else
                    \State $\tilde h^{(t)}_v \gets h_{v}^{(t,\text{flow})}$
                \EndIf
            \EndFor \label{alg:merge-end}
            \State ${h}_v^{t-1} \gets \tilde h^{(t)}_{\mathcal{V}}$ \Comment{Set up for next time step}
        \EndFor
        \LComment{Use Rainfall Forecast to predict only at targets}
        \State $\widehat{Y}_{v \in \mathcal{V_{\rho}}}^{t+\Delta t} \gets \mathrm{Predictor}(\tilde h^{(1\rightarrow T)}_{\mathcal{V}}, P^{t+\Delta t})$
        \State Compute loss at target nodes.
        \State \textbf{\textsc{Synchronize}} \Comment{Gradient sync. across trainers}
        \State Update model parameters $\Psi$
    \EndFor
    \State \textbf{return} ${\Psi}^*$
\end{algorithmic}
}
\end{algorithm}
\vspace{-1em}
\subsection{Temporal Encoding}\label{subsec:temp}
The architecture of HydroGAT is shown in Fig.~\ref{fig:architecture}. The \emph{temporal encoder} shown in blue (Line~\ref{alg:temporal-encoder}, Algorithm \ref{alg:model}) uses transformer layers to convert each node's historical precipitation and discharge into a sequence of context-rich embeddings. Applied independently along each node’s temporal dimension (while spatial relationships are preserved for the subsequent GNN), this identifies both long-range hydrological signals (e.g., seasonal baseflow trends) and short-term drivers (e.g., storm-induced runoff peaks). Unlike recurrent networks, it can attend over long horizons without vanishing gradients \cite{hochreiter1998vanishing}. While recent graph transformers~\cite{ying2021transformers, yu2020spatio, mialon2021graphit, wu2021representing} combine global attention with message passing in a single block, it scales quadratically with graph size. HydroGAT therefore separates temporal transformers from spatial GNNs, still capturing long-range dependencies while remaining computationally efficient. To better attend to hydrologically salient intervals while maintaining temporal continuity, we integrate causal masking with a \textit{precipitation-aware} attention bias.

\subsubsection{\textbf{Positional Encoding}}
Hydrological signals exhibit strong seasonal patterns, delayed runoff responses, and long-range dependencies, so it is crucial that our temporal encoder knows not only what the values are but also when they occur. To inject a sense of ``time" into each input vector, we first linearly project the raw node features at each past timestep into a higher-dimensional space using a learnable weight matrix $W_{in}\in \mathbb{R}^{F\times d}$, and then add a fixed sine-cosine positional vector ($p^t$) (inspired by \cite{vaswani2017attention}) that uniquely encodes each timestep. Formally, given a temporal sequence of node features $\mathbf{X}^{t}$, we compute for each $t = \{1, \dots, T\}$, encoding ($\mathbf{E}^t$):
\begin{align}
    \mathbf{E}^t = \mathbf{x}^t \mathbf{W}_{\text{in}} + p^t, \quad \text{where } p^t \in  \mathbb{R}^{d}
\end{align}
where $d$ is the model dimension. For e.g., by adding $p^t$, the encoder can distinguish yesterday's inputs from the 1-2 days ago even though the values themselves may be similar, improving its ability to learn both short-term and long-term relationships.

\subsubsection{\textbf{Causal Temporal Attention}} 
In hydrology, long-term and short-term patterns occur at very different time scales. A standard recurrent network treats every past input equally, but can struggle to decide which past hours are truly relevant for a particular event. In contrast, a multi-headed self-attention layer can learn from the data to focus (``attend") on exactly those past time steps, whether it was 6 or 36 hours ago, that most influence the discharge at the current timestep, while completely ignoring irrelevant periods of dry weather. At each historical timestep $t\in T$, we construct a \emph{query} vector $\mathbf{q}_t$ that asks which past hours best predict discharge at $t$. $\mathbf{q}_t$ is compared to a set of \emph{key} and \emph{value} vectors $\{\mathbf{k}_\tau,\mathbf{v}_{\tau}\}$ where each $\tau \leq t$. A causal mask ensures that no future information ($\tau>t$) is used for attention. Since rainfall older than 24 hours usually drains away from local nodes, we restrict each query to attend only to the most recent 24 hours. This is applied across every timestep in the input sequence with a sliding window of length 24. For e.g., with a 72-hour input window, each query at $t \in \{0,\dots, 71\}$ attends only to keys in range $max(0,t-23)\leq\tau \leq t$. If $d$ is the attention dimension, then for each $t$, we compute raw attention logits as below, where $-\infty$ denotes a masked position outside the sliding attention window (no attention):
\begin{align}
   \ell_{t,\tau} \;=\;
   \begin{cases}
     \displaystyle\frac{q_t \cdot k_\tau}{\sqrt{d}}\;, & \tau \in [\max(0,t-23),\,t],\\[1em]
     -\infty\;, & \text{otherwise},
   \end{cases}
\end{align}
Then we normalize via softmax over $\tau$:
\begin{align}
   \alpha_{t,\tau} \;=\; \frac{\exp(\ell_{t,\tau})}{\sum_{\tau'}\exp(\ell_{t,\tau'})}\,.
\end{align}
Finally, the updated embedding at time $t$ is
\begin{align}
   e_t \;=\; \sum_{\tau=\max(0,t-23)}^{t} \alpha_{t,\tau}\,v_\tau\,,
\end{align}
These temporal node embeddings ($e_t \in E_{seq}$) are then passed through feed-forward and residual layers, and are used as inputs to the spatial module.
 
\subsection{Gated Message Passing for Spatial Routing}\label{subsec:spatial}
Given the temporal embeddings from the temporal module for each node at time $t$, we perform spatial message passing over two edge types, $\mathcal{E_\text{F}}$ (flow) and $\mathcal{E_\text{C}}$ (catchment) edges. Each set of edges is fed to Gated Recurrent Units (GRUs) \cite{chung2014empirical} whose linear projections are replaced by graph-attention convolutions (GATs) \cite{velivckovic2017graph}. As a result, the gating parameters are learned from neighborhood messages instead of purely local feed-forward layers. 

\subsubsection{\textbf{Heterogeneous Spatial Routing}} If \(\mathcal{G}_{\text{flow}}\) and \(\mathcal{G}_{\text{catch}}\) are the subgraphs with flow and catchment edges respectively, then each edge branch (\(b \in \{\text{flow},\text{catch}\}\)) with GRU gate types, \textit{update} ($z$) and \textit{reset} ($r$) on each node $v \in V$ computes:
\begin{align}
  \mathbf{z}_{v}^{t,b} \;=\; \sigma\bigl(\mathrm{GAT}_{z}^{b}(\mathcal{G}_{b},\,\mathbf{e}^t)_v\bigr),
  \quad
  \mathbf{r}_{v}^{t,b} \;=\; \sigma\bigl(\mathrm{GAT}_{r}^{b}(\mathcal{G}_{b},\,\mathbf{e}^t)_v\bigr).
\end{align}
These gates \(\mathbf{z}_v^{t,b}\) and \(\mathbf{r}_v^{t,b}\) use the edge-wise gating-attention mechanism in GATs to allow each node in each branch adaptively filter incoming temporal messages from neighbors along flow or catchment edges (wet, active upstream neighbors receive higher attention; dry ones are suppressed). In effect, this converts hydrological context into \emph{data-driven, time-varying edge weights}, letting the graph dynamically set its edge weights at every timestep instead of relying on static coefficients. Next, each branch updates its hidden state using the GRU's reset gate:
\begin{align}
  \tilde{\mathbf{u}}_v^{t,b}
  \;=\;
  \bigl[\mathbf{e}_v^t \;\|\; \mathbf{r}_v^{t,b} \odot \mathbf{h}_v^{t-1}\bigr]
\end{align}
which allows the model to selectively forget irrelevant parts of the previous hidden state ($\mathbf{h}_v^{t-1}$) before integrating with the current input ($\mathbf{e}_v^t$) (in Fig. \ref{fig:architecture}, represented as $h_{prev}$ and $e_t$ for simplicity). Then a third GAT is applied to produce a candidate state:
\begin{align}
  \tilde{\mathbf{c}}_v^{t,b}
  \;=\;
  \tanh\Bigl(\mathrm{GAT}_{h}^{b}(\mathcal{G}_{b},\,\tilde{\mathbf{u}}^{t,b})_v\Bigr)
\end{align}
These candidates are then combined with the previous hidden state with standard GRU rule, generating the updated hidden state $\mathbf{h}_{v}^{t,b}$ for each branch \({\mathbf{h}}_v^{t,\text{flow }} \forall v \in \mathcal{V}\) and \({\mathbf{h}}_v^{t,\text{catch }} \forall v \in \mathcal{V_{\rho}}\) (Line~\ref{alg:gatflow}-\ref{alg:gatcatch}):
\begin{align}
  \mathbf{h}_{v}^{t,b}
  \;=\;
  \bigl(1 - \mathbf{z}_v^{t,b}\bigr)\!\odot\,\mathbf{h}_v^{t-1}
  \;+\;
  \mathbf{z}_v^{t,b} \!\odot\,\tilde{\mathbf{c}}_v^{t,b}
\end{align}
\subsubsection{\textbf{Gated Learnable Parameter $(\boldsymbol{\alpha})$}}
To merge the outputs of two spatial modules, we introduce a learnable parameter vector \(\boldsymbol{\alpha}\in\mathbb{R}^H\) (one scalar per attention head $H$) which is passed through a sigmoid to lie in \((0,1)\) (Line~\ref{alg:alphainit}). The resulting $\alpha$ is broadcast across the hidden dimension of each head and applied only to target nodes $v \in V_{\rho}$. Formally, for each target node at each timestep \(t\), we compute the fused hidden state with (Line~\ref{alg:merge-start}-\ref{alg:merge-end}):
\begin{align}
  \tilde{\mathbf{h}}_v^t
  \;=\;
  \alpha \odot {\mathbf{h}}_v^{t,\text{flow}}
  \;+\;
  \bigl(1 - \alpha\bigr) \odot {\mathbf{h}}_v^{t,\text{catch}}
\end{align}
Our spatial module provides two key benefits: (i) \textit{parallel edge-type routing} via separate flow and catchment edges combined by a learnable weight $\alpha$  that dynamically balances these two sources of information under varying conditions (e.g., a swollen tributary vs.\ dry overland flow); and (ii) \textit{gated attention} that selectively propagates information from active upstream nodes, emphasizing active contributors while suppressing inactive or dry neighbors. This improves the model's ability to route water through the network. 

\subsection{Prediction using Forecasted Rainfall}
To forecast discharge over the next $\Delta t$ steps, the model incorporates external real-time precipitation forecasts. This does not violate causality, as both the temporal and spatial modules use only past observed data. The future rainfall forecast is transformed separately by a lightweight convolutional layer that learns a local representation of the anticipated precipitation. We then concatenate the transformed rainfall with the fused hidden state $\tilde{\mathbf{h}}_v^t$ along the feature dimension. The combined representation is passed through a learnable convolutional fusion block to generate the final discharge prediction at each target node for the next $\Delta t$ time steps.
\subsection{Distributed Training Pipeline}\label{subsec:dist}
\begin{figure}[h]
  \centering
  \includegraphics[width=\columnwidth]{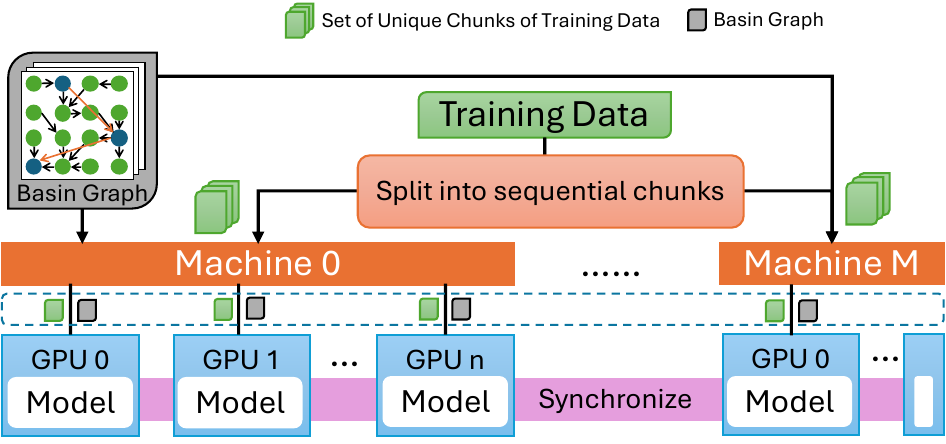}
  \caption{Proposed distributed training pipeline. Temporal data are split into sequential chunks. The basin graph is replicated across machines. Each GPU trains a model replica on its assigned chunk, with periodic synchronization.}\vspace{-1em}
  \label{fig:dist}
\end{figure}
To support high-resolution spatiotemporal modeling over large basins and long temporal sequences, we implement a distributed training strategy (Fig.~\ref{fig:dist}) using PyTorch's Distributed Data Parallel (DDP)\footnote{\url{https://docs.pytorch.org/docs/stable/notes/ddp.html}} framework. This approach allows the model to scale across multiple machines and GPUs in HPC environments. As training begins, one training process is spawned per GPU. Each process holds the complete basin graph (shown in grey) and a full replica of the model parameters that are initialized identically. Since the graph itself fits comfortably in the CPU memory of a single machine, even at high resolution, we avoid partitioning it across machines. Preserving temporal integrity and causal dependencies is critical in hydrological forecasting (sequential dependencies govern runoff generation and flow routing). Therefore, we adopt full-batch training \cite{mostafa2021sequential,yuan2023comprehensive} with a custom \textit{sequential distributed sampler} that assigns each process a temporally contiguous, non-overlapping segment (chunks) of the dataset. Training samples are extracted via a fixed sliding window of size $(T_{\text{in}} + T_{\text{out}})$ that slides one hour per step, producing input-output sequence pairs where precipitation inputs are returned for all nodes, but discharge, both as input and labels are returned only for target nodes. The sampler preserves the hydrological sequence while enabling efficient parallel training. Gradients are synchronized via \texttt{AllReduce}, ensuring consistent parameter updates across model replicas. Since all replicas remain synchronized throughout training, any one can be saved as the final model.

During inference, we use the same sampler to load evaluation data into the model. The model generates overlapping predictions from each evaluation window, which are then stitched into a continuous predicted discharge sequence. Since the windows overlap, the same timestep $t$ may appear in multiple predicted sequences. To obtain a final, continuous prediction, we aggregate these overlapping predictions by averaging them: $\hat{\mathbf{Y}}[t] = \frac{1}{|\mathcal{W}_t|} \sum_{i \in \mathcal{W}_t} \hat{\mathbf{Y}}_i[t]$, where \( \mathcal{W}_t \) is the set of all windows covering timestep \(t\) and $\hat{\mathbf{Y}}_i[t]$ is the prediction for timestep $t$ from window $i$. This mitigates the higher variance that arises at the extreme lead times of each 72-hour forecast (e.g., the last few hours), resulting in a smoother and more reliable continuous discharge sequence.
\section{Results}\label{sec:results}
We introduce the study region in \S\ref{subsubsec:studyregion}, the software/hardware platform in \S\ref{subsubsec:platform}, the baselines in \S\ref{subsubsec:baselines}, and the evaluation metrics in \S\ref{subsubsec:metrics}. All models are trained and tested using identical preprocessing steps and model hyperparameters to ensure fair and reproducible benchmarking across heterogeneous river basins.

\subsection{Experimental Setup}
\subsubsection{\textbf{Study Region}}\label{subsubsec:studyregion}
\begin{figure}[ht]
  \centering
  \includegraphics[width=0.4\textwidth]{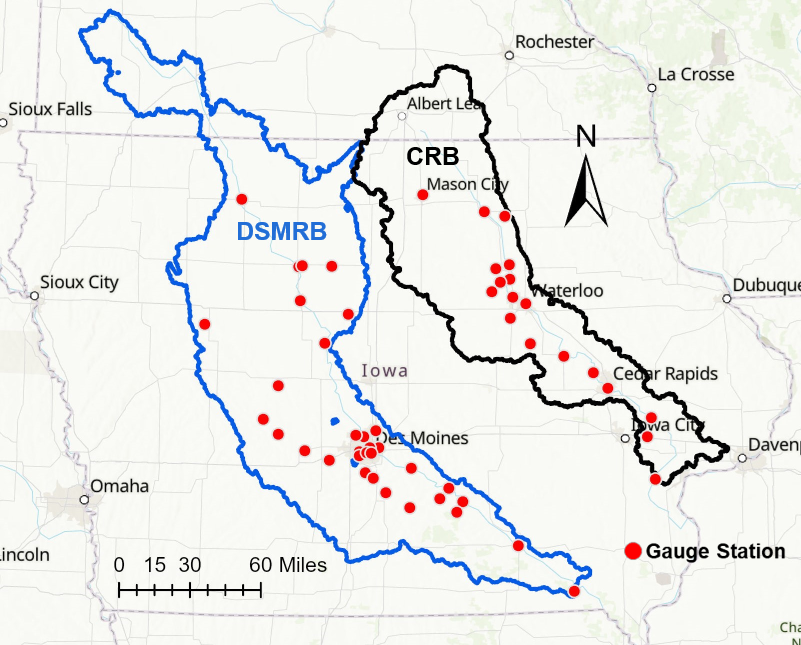}
    \caption{The locations of the Cedar River Basin (CRB) and Des Moines River Basin (DSMRB) in Midwest-USA. Red dots indicate the gauge stations, which serve as "target" nodes (see \S\ref{subsubsec:nodes-definition}) in our framework.}\vspace{-1em}
  \label{fig:basin-location}
\end{figure}

We evaluate HydroGAT on two major river basins in the U.S. Midwest shown in Fig.~\ref{fig:basin-location}: the \textit{Des Moines River Basin} (DSMRB) and the \textit{Cedar River Basin} (CRB). These basins differ in land use, drainage density, and elevation, providing a diverse testbed for flood forecasting. DSMRB spans a larger area with an extensive tributary network, while the CRB is smaller but features more local hydrological variability due to steeper slopes and more localized runoff. Hydrological analysis indicates that sub-basins within the DSMRB tend to exhibit wetter conditions and higher discharge levels compared to those in the CRB. 

Each basin is discretized into a 4km$\times$4km grid covering both land and river pixels. Directed flow edges are calculated using the 1 arc-second Digital Elevation Model (DEM) from the USGS\footnote{United States Geological Survey} 3D Elevation Program~\cite{usgs2021dem}. We clipped the DEM to the study region, resampled it to 4 km resolution, and processed it using the ArcGIS Pro "Fill" and "Flow Direction" tools. First, the Fill tool removes depressions to ensure hydrologic connectivity; then the Flow Direction tool computes the pixel-level flow directions. Each node has hourly precipitation data from 2012-2018 collected from the NCEP/EMC Stage IV product~\cite{du2011stageiv}. Upstream-downstream catchment relationships and hourly discharge observations at target gauge stations aggregated from USGS sensor stations are obtained from WaterBench~\cite{demir2022waterbench}. More details about the basin graphs are provided in Table~\ref {tab:basin_summary}.
\begin{table}[ht]
  \centering
  \small
  \caption{Summary of Cedar River Basin (CRB) and Des Moines River Basin (DSMRB).}
  \label{tab:basin_summary}
  \begin{tabular}{l c c c c c}
    \toprule
    \textbf{\shortstack{Basin\\ Name}} 
      & \textbf{\shortstack{Area\\(in $miles^{2}$)}}
      & \textbf{\#Nodes }
      & \textbf{\shortstack{\#Edges\\(Flow)}}
      & \textbf{\shortstack{\#Edges\\(Catchment)}} 
      & \textbf{\shortstack{\#Gauge\\Stations}} \\
    \midrule
    CRB         & 7980  & 1288 &  1247     &   17    & 18 \\
    DSMRB       & 14014 & 2226 &  2157     &   32    & 33 \\
    \bottomrule
  \end{tabular}
\end{table}

To address missing hourly discharge values in the WaterBench dataset (due to station/system failures, connectivity outages, and winter ice cover), we implemented a two-step imputation strategy based on station location and flow behavior. First, for the most downstream station, where missing values occurred only sporadically due to equipment malfunction or river freezing, we applied linear interpolation to estimate the missing values. Given the gradual variability of streamflow at hourly resolution, this provided reliable estimates without distorting temporal trends. Second, for upstream stations, we leveraged the hydrological continuity between upstream and downstream sites by fitting station-specific linear regression models. Each model parameterized the upstream discharge \(Q_{\text{up}}\) as a linear function of its hydrologically connected downstream counterpart \(Q_{\text{down}}\): \(Q_{\text{up}} = a + b \cdot Q_{\text{down}}\), where \(a\) and \(b\) are regression coefficients estimated from concurrent, non-missing observations. These models were then used to estimate missing upstream values, while original observations were retained to preserve data integrity.
\begin{table*}[!t]
\centering
\small
\caption{Basin-level NSE evaluated across different baselines using 72 input hours to predict the next 72 output hours against several evaluation metrics ($\downarrow$ means lower is better and $\uparrow$ means higher is better). All baselines were trained using our distributed pipeline (see \S\ref{subsec:dist}) using 4 machines (16 GPUs, 4 GPUs/machine). Best shown in bold and second-best underlined.}
\label{tab:arch_comparison}
\begin{tabular}{@{}l  *{6}{c}  *{6}{c}@{}}
\toprule
\textbf{Architecture}
  & \multicolumn{6}{c}{\textbf{Cedar River Basin}} 
  & \multicolumn{6}{c}{\textbf{Des Moines River Basin}}
\\
\cmidrule(lr){2-7} \cmidrule(lr){8-13}
  & NSE$\uparrow$ & KGE$\uparrow$ & NRMSE$\downarrow$ & NMAE$\downarrow$ 
  & MAPE$\downarrow$ & PBIAS(\%) 
  & NSE$\uparrow$ & KGE$\uparrow$ & NRMSE$\downarrow$ & NMAE$\downarrow$ 
  & MAPE$\downarrow$ & PBIAS(\%) 
\\
\midrule
DCRNN          & \underline{0.95} & \textbf{0.96} & \underline{37.15\%} &                                \underline{14.48\%} & \underline{15.35\%} & \textbf{-0.96\%}  
               & \textbf{0.97} & 0.90 &  26.81\% & 11.74\% & \underline{21.85\%} &  6.81\%  \\
GraphWaveNet   & 0.92 & \underline{0.94} & 44.05\% & 17.53\% & 15.36\% & -1.29\% 
               & \textbf{0.97} & \underline{0.92} & \textbf{25.19\%} & \textbf{10.68\%} & 24.30\% & \underline{4.61\%}  \\
RGCN           & 0.90 & 0.89 & 48.93\% & 18.06\% & 18.59\% &  2.91\% 
               & \textbf{0.97} & \textbf{0.97}& \underline{26.35\%} & 11.71\% & 25.58\% & \textbf{0.60\%} \\
GCRNN          & 0.86 & 0.84 & 59.55\% & 25.26\% & 26.57\% &  5.90\% 
               & 0.84 & 0.74 & 64.44\% & 27.53\% & 39.08\% & 16.91\% \\
STGCN-WAVE     & 0.76 & 0.82 & 77.62\% & 31.61\% & 31.20\% &  1.81\% 
               & \underline{0.90} & 0.86 & 49.44\%  & 21.30\%  & 45.13\%  & 8.65\% \\
\midrule
\textbf{HydroGAT (Ours)}  & \textbf{0.97} & \textbf{0.96} & \textbf{30.15\%} & \textbf{12.14\%} & \textbf{13.28\%} & \underline{-1.10\%}  
               & \textbf{0.97} & \underline{0.92} &  26.46\% & \underline{10.71\%} & \textbf{19.31\%}  & 4.70\% \\
\bottomrule
\end{tabular}
\label{tab:baseline-comparison}
\end{table*}

\subsubsection{\textbf{Normalization of Features and Labels}}\label{subsec:normalization}
\begin{figure}[h]
  \centering
  \begin{subfigure}[b]{0.49\columnwidth}
    \includegraphics[width=\textwidth,keepaspectratio]{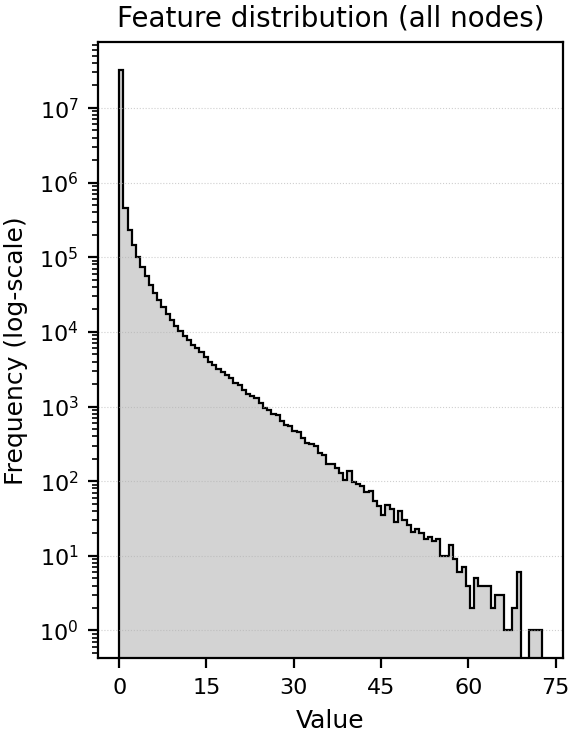}
  \caption{}
  \label{fig:feature_distribution}
  \end{subfigure}
  \hfill
  \begin{subfigure}[b]{0.49\columnwidth}
    \includegraphics[width=\textwidth,keepaspectratio]{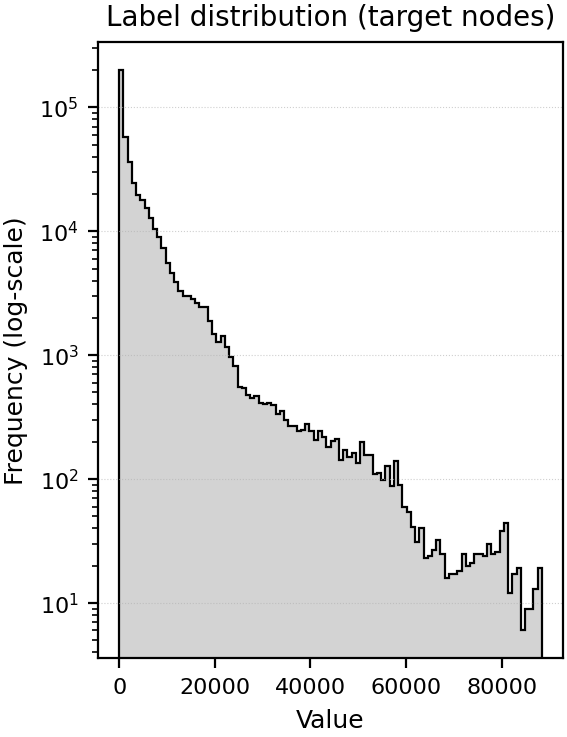}
  \caption{}
  \label{fig:label_distribution}
  \end{subfigure}
  \caption{Distributions of raw precipitation (features) (a) and discharge labels (b) prior to normalization. Both exhibit heavy-tailed behavior and wide dynamic ranges, motivating the use of \(\boldsymbol{\log_1p}\) and min-max transformations.}\vspace{-1em}
  \label{fig:feature_label_distribution}
\end{figure}
The features and labels have a broad range of values (Figure~\ref {fig:feature_label_distribution}). To mitigate skewness in hydrological variables and stabilize model training, we apply a two-step transformation to both input features and discharge labels: a \(\log_1p\) transformation followed by min-max normalization. This process maps all non-negative values to the \([0,1]\) interval while compressing extreme values. Given a raw time series \(Z \in \mathbb{R}^{n \times T}\) (e.g., precipitation or discharge), we apply a non-negativity correction and logarithmic transformation ($Z^{(\log)} = \log(1 + \max(Z, 0))$) before normalizing using min-max scaling. 
\subsubsection{\textbf{Platform and Model Hyperparameters}}\label{subsubsec:platform} Experiments were run on NERSC Perlmutter supercomputer (1,792 nodes, dual 64-core AMD EPYC 7763 CPUs, 256GB RAM; 4$\times$ NVIDIA A100 GPUs; HPE Slingshot 11 interconnect \cite{yang2020accelerate}). Model is implemented in PyTorch v2.1 with mixed precision, CUDA v11.8, and DGL v1.1. We train for 100 epochs with AdamW optimizer~\cite{loshchilov2017decoupled} (LR=$0.01$, weight decay=$1\mathrm{e}{-4}$, patience=5), batch size=8, 72 hours input and 72 hours output windows, 32 hidden features, 2 attention heads/module, and 0.1 dropout. Data span May-September from 2012–2018, split into 4 training, 1 validation, and 2 test years.
\subsubsection{\textbf{Baselines}} \label{subsubsec:baselines}
Many recent advances in spatiotemporal GNNs have been driven by research in traffic forecasting, a domain structurally similar to hydrology, where signals (e.g., vehicle flow or streamflow) evolve over sparse, directed graphs with strong spatiotemporal dependencies. Foundational models like DCRNN~\cite{li2018diffusion} and STGCN~\cite{yu2017spatio} introduced diffusion-based graph traversal and dilated temporal convolutions, which have since been adapted to hydrology~\cite{sun2021explore}. We therefore choose five state-of-the-art spatiotemporal architectures widely used in graph-based modeling in both traffic prediction and hydrology. These include the Diffusion Convolutional Recurrent Neural Network (DCRNN)~\cite{li2018diffusion}, GraphWaveNet~\cite{sun2021explore}, Recurrent Graph Convolutional Network (RGCN) ~\cite{sit2021shorttermhourlystreamflowprediction}, STGCN-WAVE~\cite{yu2017spatio, sun2021explore} (STGCN's graph convolution layers combined with WaveNet-style dilated temporal filter), and a Graph Convolutional Recurrent Neural Network (GCRNN)~\cite{zanfei2022graph}. GraphWaveNet and STGCN-WAVE adopt fully convolutional temporal modules and do not rely on recurrent units, distinguishing them from the other recurrent-based baselines. All baselines were implemented or adapted to use our basin graphs, input-output window, and distributed training \& evaluation pipelines, the same setup as our proposed model, to ensure a fair comparison. This adaptation yielded substantial gains over the original baseline designs. 

\subsubsection{\textbf{Evaluation Metrics}}\label{subsubsec:metrics}
We evaluate model performance using standard hydrological metrics. Nash-Sutcliffe Efficiency (NSE)~\cite{waseem2017review} $(-\infty, 1]$, compares predictions to the observed mean ($>0.5$ are acceptable). Percent Bias (PBIAS)~\cite{waseem2017review} indicates bias, where $>0$ denotes overestimation and $<0$ underestimation. Kling-Gupta Efficiency (KGE)~\cite{waseem2017review} $(-\infty, 1]$, combines correlation, bias, and variability (1 is ideal). We also report scale invariant metrics like Normalized Root Mean Squared Error (NRMSE), Normalized Mean Absolute Error (NMAE), and Mean Absolute Percentage Error (MAPE), both in $[0, \infty)$, where lower is better. 

\subsection{Comparison with Baselines}
Table \ref{tab:baseline-comparison} compares six spatiotemporal baselines with HydroGAT for predicting hourly discharge on two basins using a 72-hour input-output window. HydroGAT consistently outperforms or matches the best baselines across all metrics. In the CRB, it achieves the highest NSE (0.97), which represents the overall accuracy of the model forecast throughout the region and forecast window. It ties with DCRNN for best KGE (0.96), and outperforms all baselines in NRMSE, NMAE, MAPE, and PBIAS. In DSMRB, it leads with NSE=0.97 and KGE=0.92, slightly improving over DCRNN. Across both basins, our method offers the best trade-off between accuracy (NSE/KGE) and error, demonstrating its ability to more faithfully reproduce both the timing and magnitude of observed discharge compared to existing spatiotemporal networks and recurrent‐based methods. 
\subsubsection{\textbf{Basin performance by lead time}}
\begin{figure}[h]
  \centering
  \includegraphics[width=\columnwidth,keepaspectratio]{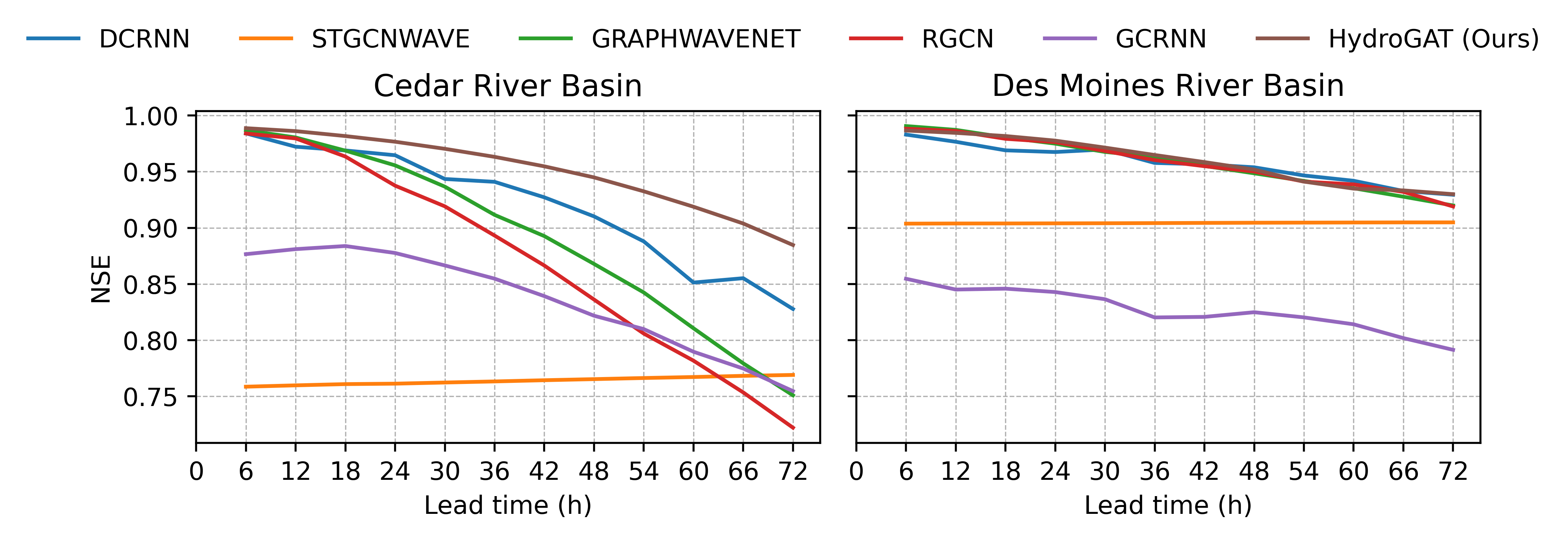}
  \caption{Basin-level NSE of hourly discharge prediction along the gradient of lead time, 6 to 72 hours at 6-hour intervals. [\emph{Higher} is better]}\vspace{-1.8em}
  \label{fig:horizon-nse}
\end{figure}

Fig. \ref{fig:horizon-nse} complements Table~\ref{tab:baseline-comparison} by showing performance across varying lead times from 6 to 72 hours at 6-hour intervals. In CRB, our model maintains NSE>0.90 until 66 hours, while DCRNN drops below 0.90 after 50 hours and GraphWaveNet after 24 hours. In Des Moines, several baselines perform competitively early on. However, after about 66 hours, their NSEs dip more steeply, whereas our model preserves a small but consistent margin and finishes the 72-hour horizon with the highest NSE. This performance is beneficial for guiding regional-scale decision-making.

\subsection{Sub-basin Performance}
\begin{figure}[h]
  \centering
  \begin{subfigure}[b]{0.49\columnwidth}
    \includegraphics[width=\textwidth,keepaspectratio]{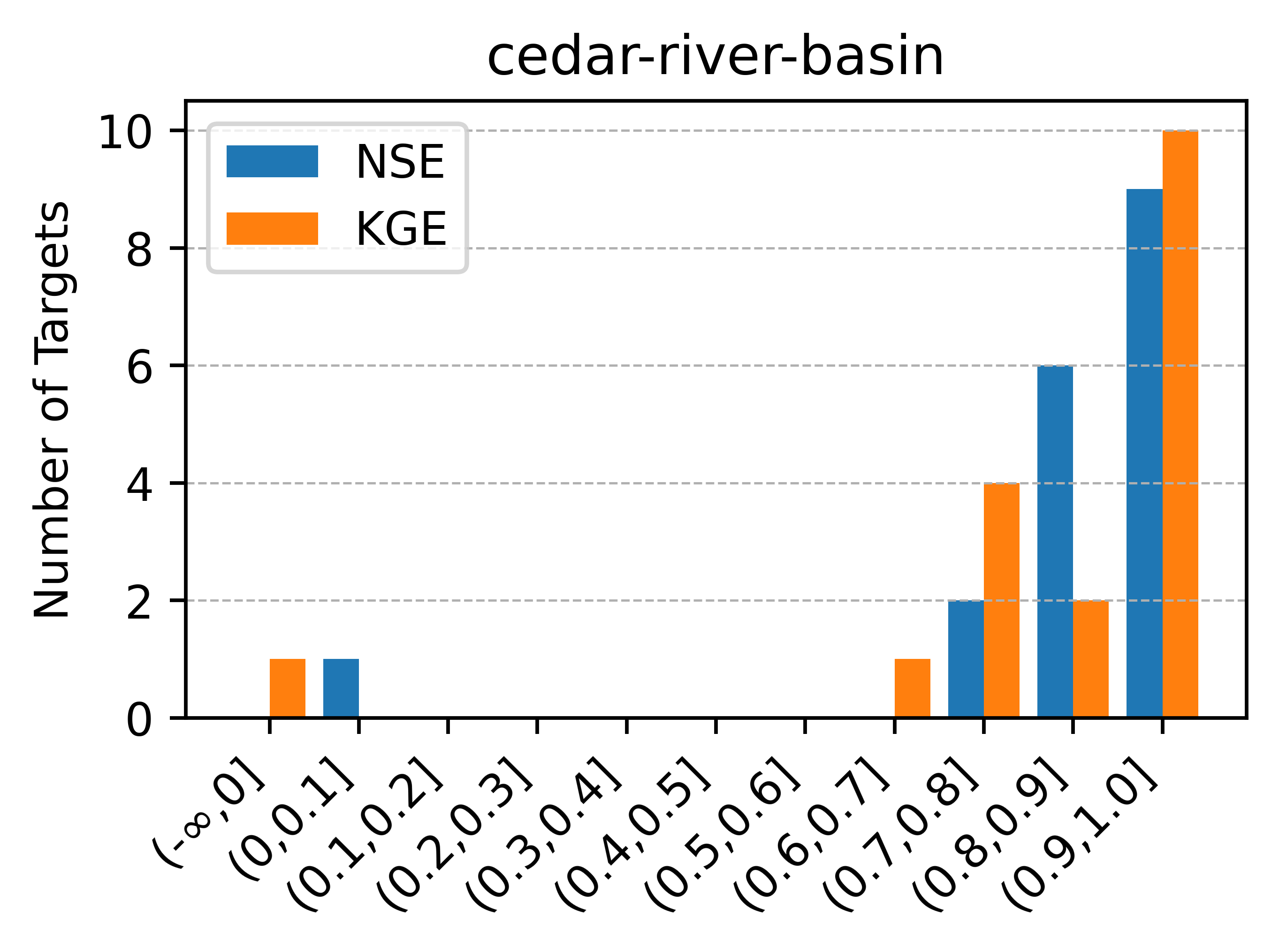}
    \caption{}
    \label{fig:cedar-per-target-nse}
  \end{subfigure}
  \hfill
  \begin{subfigure}[b]{0.49\columnwidth}
    \includegraphics[width=\textwidth,keepaspectratio]{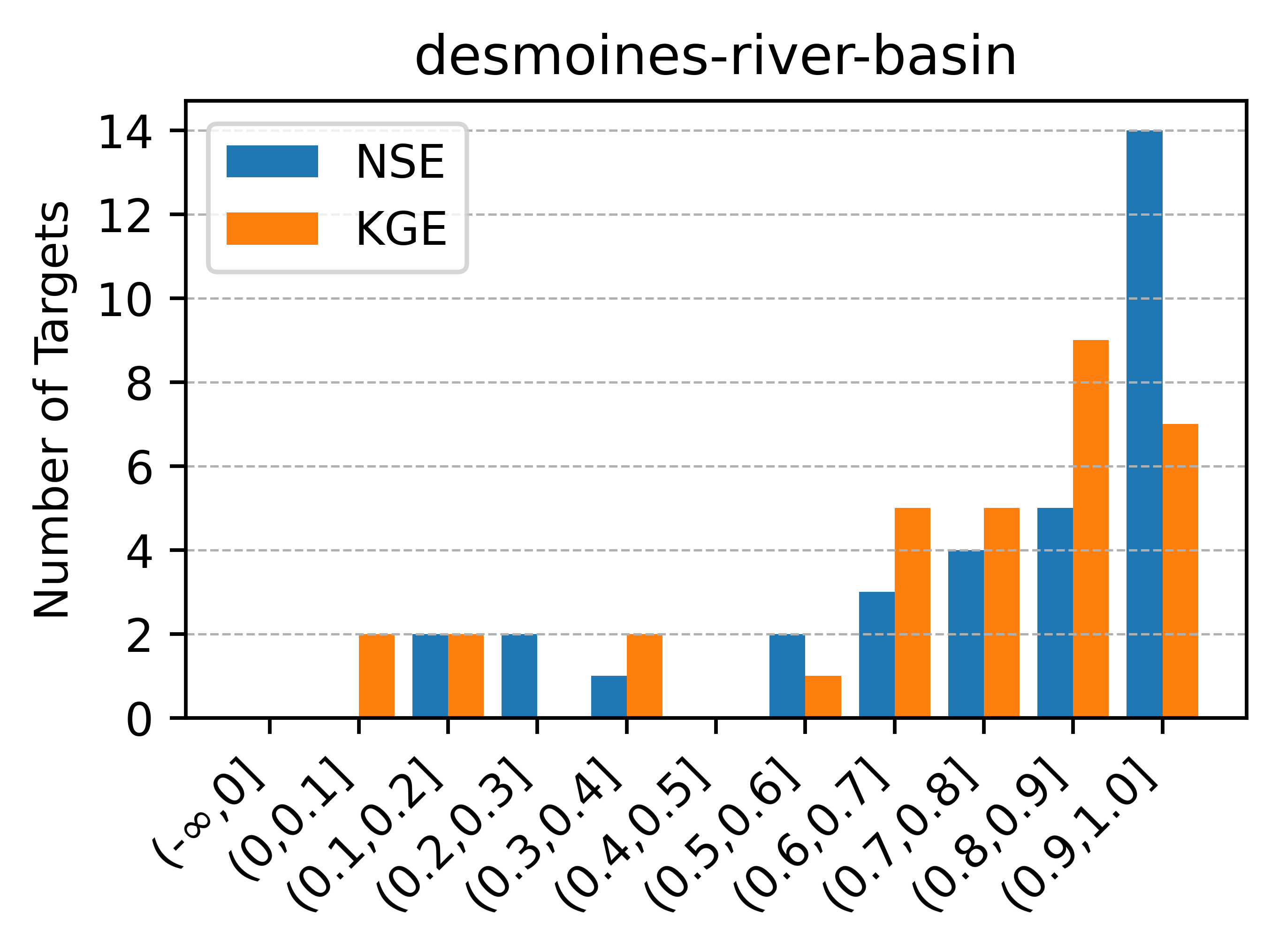}
    \caption{}
    \label{fig:dsm-per-target-nse}
  \end{subfigure}
  \caption{NSE/KGE score distributions for hourly discharge prediction. Each bar shows the number of target gauge stations that achieved NSE/KGE scores within each performance interval over a 72-hour forecast horizon. [\emph{Higher} is better.]}
  \label{fig:nse_kge_distribution}
\end{figure}
\begin{figure}[b] 
  \centering
  \includegraphics[width=\columnwidth,keepaspectratio]{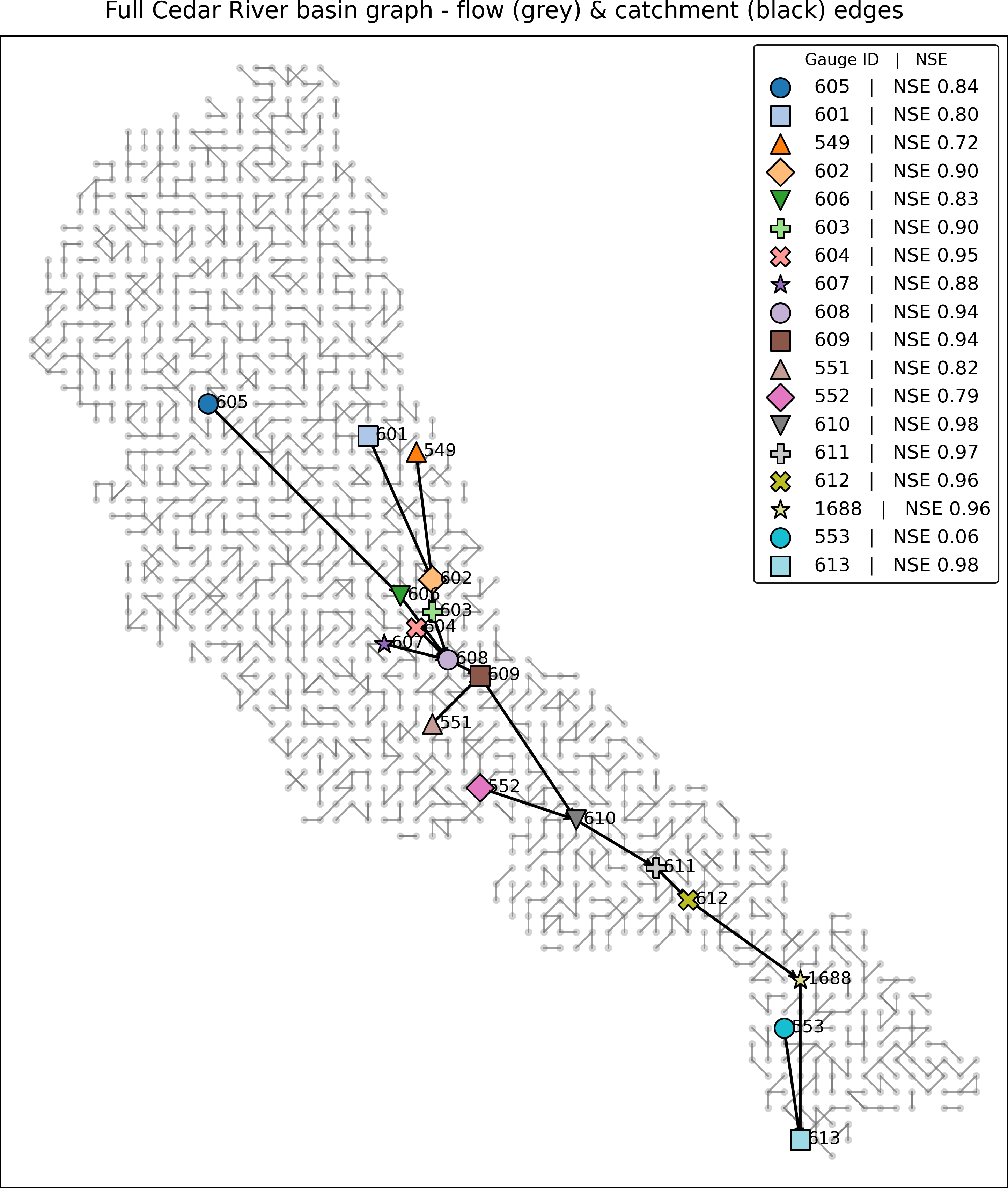}
  \caption{Cedar River basin graph with flow (grey) and catchment (black) edges.
Each colored marker represents a gauged station, annotated with its gauge ID and the aggregated NSE score for hourly discharge prediction over a 72-hour output window, evaluated across 2017-2018. [\emph{Higher} is better]}\vspace{-1em}
  \label{fig:cedar_river_full_graph}
\end{figure}
\begin{figure}[h] 
  \centering
  \includegraphics[width=\columnwidth,keepaspectratio]{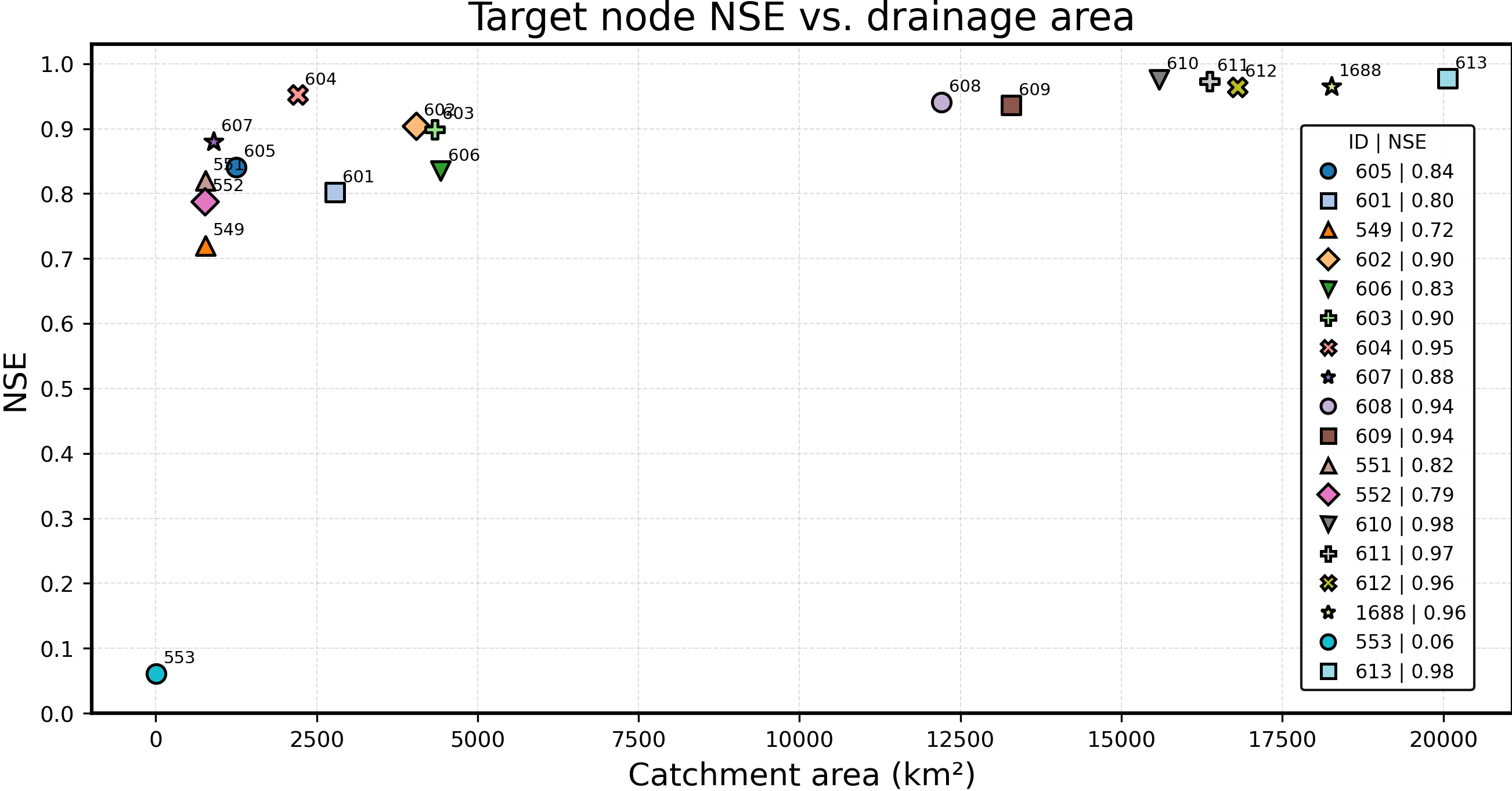}
  \caption{NSE per target node vs.\ drainage area in Cedar Basin. [\emph{Higher} is better]}\vspace{-1em}
  \label{fig:nse_vs_area}
\end{figure}
\subsubsection{\textbf{Aggregated Sub-basin Performance}} 
\begin{figure}[t]
  \centering
    \includegraphics[width=0.5\textwidth,keepaspectratio]{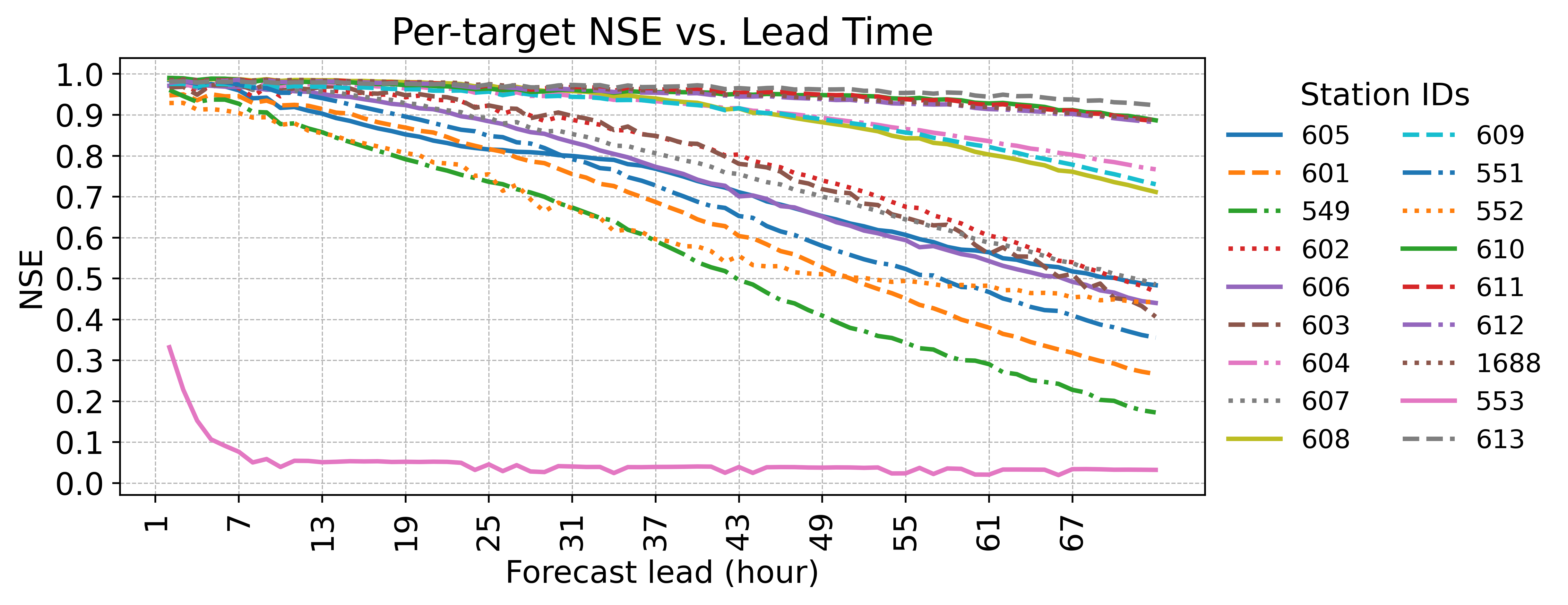}
    \caption{NSE computed per gauge station in Cedar River Basin across 72 hours. [\emph{Higher} is better]}\vspace{-1.8em}
    \label{fig:nse_per_target_per_hour}
\end{figure}

Fig.~\ref{fig:nse_kge_distribution} summarizes the distribution of NSE and KGE scores (aggregated over a 72-hour horizon) across target stations (each represents a sub-basin) in both basins. In CRB (Fig.~\ref{fig:cedar_river_full_graph}), the model achieves strong performance at most gauge stations, with only 1 station with NSE<0.5. 10 out of 18 sub-basins achieve NSE $\geq 0.90$ despite variation in rainfall, sub-basin area, and topography. Low-performing stations such as 549, 601, and 605 (NSEs 0.72, 0.80, 0.84, respectively), which are headwater sub-basins, illustrate limitations at nodes without upstream discharge information, limiting GNN message passing. As a result, prediction is highly dependent on local rainfall accuracy, making it sensitive to errors. In contrast, station 613 at the basin outlet achieves NSE=0.98, benefiting from upstream information aggregation through dual-edge message passing. The most extreme outlier station, 553, records an average NSE of just 0.06. As shown in Fig.~\ref{fig:nse_vs_area}, this gauge drains the smallest catchment in the network ($\sim$6\,km$^2$), making it highly sensitive to localized rainfall errors. Its isolated position and lack of upstream context limit the benefits of spatial aggregation, and its small drainage area results in highly variable discharge dynamics that are difficult to capture from coarse-scale input features. Similarly, all low performing stations in the DSMRB are also associated with small catchment areas ($\sim$153-860 $km^{2})$. This aligns with established challenges in modeling small catchments, which are more sensitive to localized rainfall, exhibit heterogeneous hydrological responses, and require even higher resolution for accurate flood prediction. The model learns to ignore these outliers as will be discussed in \S\ref{subsubsec:spatial-attention}.
We also investigate how performance decays at each station across 72 lead hours in CRB (Fig. \ref{fig:nse_per_target_per_hour}). The outlet station 613 shows the most stable performance, maintaining NSE>0.90 over the full 72-hour horizon. In contrast, 553 (outlier) drops to near-zero NSE. In addition, the headwater subbasins show an overall steeper decline. \vspace{-1em}
\begin{figure}[h] 
  \centering
  \includegraphics[width=\columnwidth,keepaspectratio]{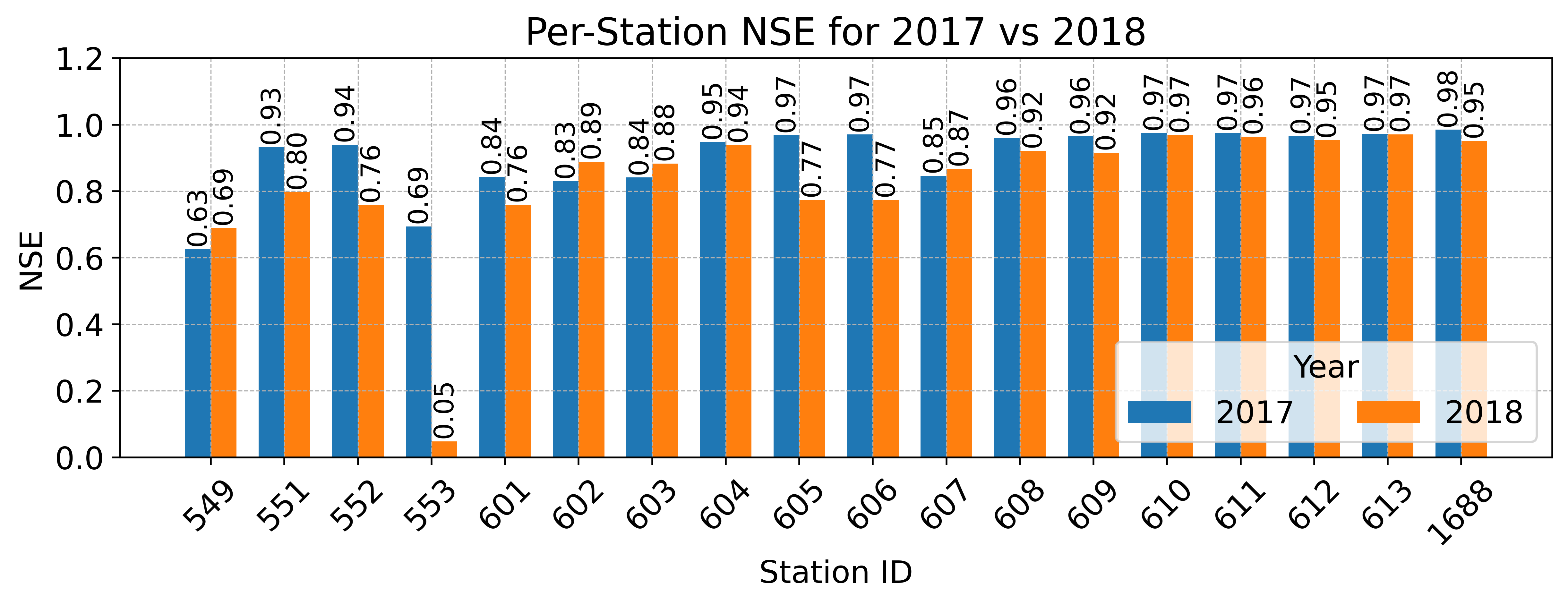}
  \caption{Annual NSE per station in the Cedar River Basin: 2017 (blue) vs 2018 (orange).  [\emph{Higher} is better]}\vspace{-1em}
  \label{fig:nse_per_year_per_target}
\end{figure}

\subsubsection{\textbf{Annual Sub-basin Performance}} Fig.~\ref{fig:nse_per_year_per_target} compares per-station NSE across 2017 and 2018 in CRB. Most gauges maintain high NSE (>0.80) in both years, reflecting stable model performance. Mid-basin stations (e.g., 604, 610–613) show consistent NSEs (0.94–0.96), suggesting that the dominant hydrological patterns did not change between years. The largest change occurs at station 553, where NSE drops from 0.69 (2017) to 0.05 (2018), highlighting the vulnerability of small catchments.
\subsection{Ablation Study}
We conducted ablation studies using shorter training runs to efficiently evaluate the impact of each architectural component in our model, isolating each module's contribution to overall performance.

\subsubsection{\textbf{Edge Weights}} We evaluated distance-based edge weights ~\cite{sarkar2023,yang2023runoff,sit2021shorttermhourlystreamflowprediction} to reflect proximity-based influence but found no performance gain. GAT layers already learn edge-specific attention weights dynamically, multiplying them by fixed, distance-based scalars imposes a static rescaling that hinders learning, especially for long-distance edges receiving smaller gradients. Furthermore, distance-to-river is a weak proxy of hydrological relevance and varies only slightly after normalization (0.05–1), adding noise rather than as a useful prior. In contrast, attention weights adaptively learn proximity when relevant, without being constrained by fixed assumptions.

\subsubsection{\textbf{Transformer Encoder vs. Naive Multiheaded Attention}}
We replaced the transformer-based encoder with a naïve multi-headed attention (MHA) that computed attention weights across timesteps without positional encoding, layer normalization, or feedforward layers. This degraded performance significantly by >10 NSE percentage points. The absence of positional encoding meant that the model could not distinguish between temporally ordered inputs, limiting its ability to capture delayed cause-and-effect relationships essential in hydrology. Unlike our temporal encoder, which incorporates a sequence of operations that promote temporal continuity and depth-wise abstraction, the naïve MHA lacked the architectural components required to model long-range dependencies and contextual hierarchies.

\subsubsection{\textbf{Spatial Attention in GNNs}}
Replacing the attention in the spatial module (the GAT layers inside the GRUs, as shown in Fig. \ref{fig:architecture}) with a Chebyshev graph convolution (ChebConv) led to poor performance. Unlike GAT, which learns edge‐specific weights based on node features and hydrological context, ChebConv applies a fixed spectral filter that cannot adaptively prioritize active upstream nodes. Its global filtering also overlooks local topographic and edge-type variations. Empirically, this led to 5-30 percentage point drops in NSE, highlighting the importance of attention-based routing for capturing dynamic hydrological dependencies.

\subsubsection{\textbf{Using Rainfall Forecast}}\label{subsub:abal-forecast}
\begin{figure}[h] 
  \centering
  \includegraphics[width=\columnwidth,keepaspectratio]{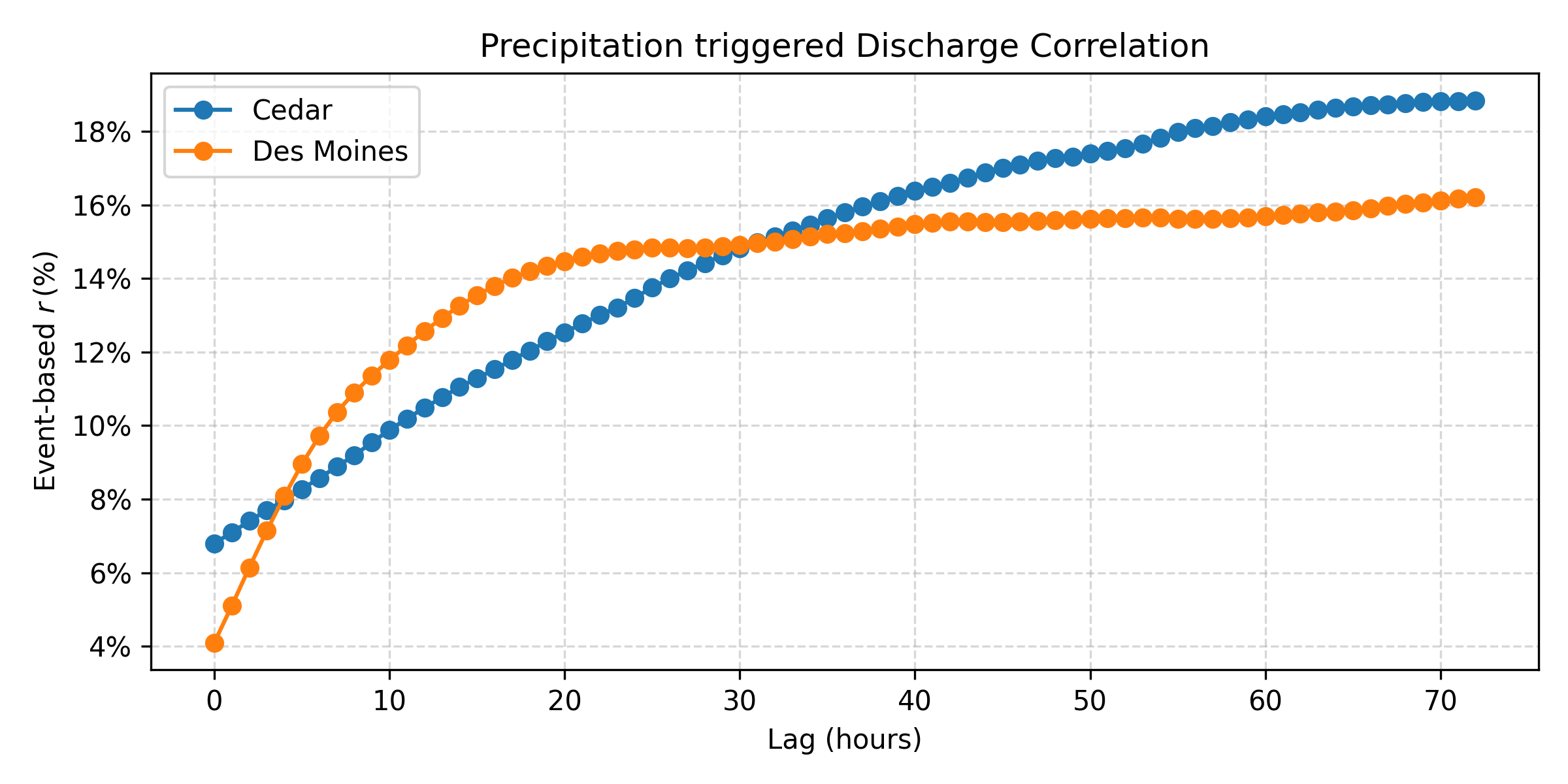}
  \caption{Event-based precipitation-discharge correlation as a function of lag $\tau$ (0-72 h) for the Cedar (blue) and Des Moines (orange) basins. At each $\tau$, the plot shows the Pearson correlation (r) between rainfall at time $t$ and discharge at $\tau+t$, computed only across rain events}\vspace{-1em}
  \label{fig:precip-corr}
\end{figure}
To quantify the benefit of conditioning the prediction on anticipated precipitation, we removed the future precipitation as input. In DSMRB, this led to a 1 percentage point drop in mean NSE, >3-point drop in KGE, increased error metrics, and a PBIAS reversal (–5.2\% to +6.1\%). CRB, however, showed no significant change. As shown in Fig. \ref{fig:precip-corr}, DSMRB shows slightly higher event-based precipitation to discharge correlation at short lags (15–16\% within 24h) than CRB, indicating that rainfall events tend to trigger observable discharge within a day. Hence, knowing tomorrow's forecast improves flood peak prediction, and removing that degrades performance. CRB shows only a modest immediate correlation ($\approx$7\% at 0 lag) and a gradual rise to ~18\% only after several days, showing slower routing across the region. In such a system, future rainfall contributes little beyond what is already encoded using the past inputs, so dropping the future branch has essentially no impact on prediction accuracy.
\begin{figure}[h] 
  \centering
  \vspace{-1.5em}
  \includegraphics[width=\columnwidth]{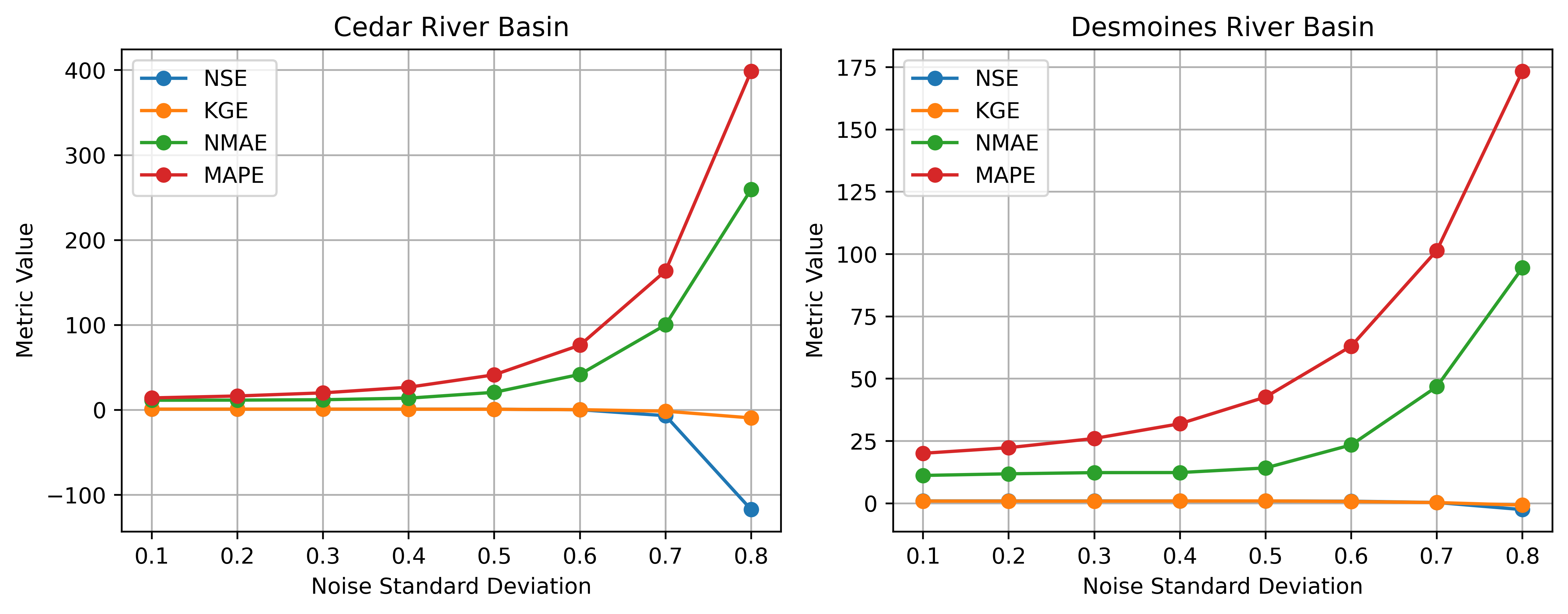}
  \caption{Sensitivity of model to Gaussian noise (STD=0.1-0.8) in rainfall forecasts.  [NSE/KGE: \emph{Higher} is better; NMAE/MAPE: \emph{Lower} is better]}\vspace{-1.5em}
  \label{fig:sensitivity}
\end{figure}

\noindent \textbf{Sensitivity.} We performed a sensitivity analysis by adding Gaussian noise (STD=0.1-0.8) to the rainfall forecast during inference and evaluated NSE, KGE, NMAE, and MAPE (Fig.~\ref{fig:sensitivity}). Both basins tolerate small forecast errors up to STD$\leq0.3$. Beyond STD=0.4, CRB collapses rapidly, whereas DSMRB remains more resilient comparatively. This difference occurs because the model, trained on clean forecasts, continues to rely on the noisy input. In CRB, the slow hydrologic response causes forecast errors to accumulate over time, leading to larger performance drops. DSMRB’s quick response limits error accumulation. Ablating the forecast branch entirely (\S\ref{subsub:abal-forecast}) caused little harm in CRB, showing that missing information is less damaging than misinformation.

\subsubsection{\textbf{Removing Catchment Edges}}
To isolate the effect of catchment edges, we evaluated HydroGAT using only flow edges (single spatial module), removing upstream-downstream connections. This resulted in a 5 percentage point drop in basin-level NSE in DSMRB and a 15-point drop in CRB. The flow-only model performs better in DSMRB (NSE=0.90) than CRB (NSE=0.74), suggesting that flow alone captures more spatial dependencies in the larger, denser DSMRB. The smaller and sparser CRB benefits more from the catchment edges, which compensate for missing long-range spatial dependencies not well represented by flow edges alone.

\subsubsection{\textbf{Learnable parameter $\boldsymbol{\alpha}$ vs. MLP}}
We replaced the $\alpha$-based fusion of the two spatial modules (see \S\ref{subsec:spatial}) with a target node-specific Multi Layer Perceptron (MLP), motivated by the possibility that the optimal relation between the flow and the catchment branches may not always be linear. However, across both basins, the learnable scalar parameter $\alpha$, which linearly applies weights to the flow and the catchment embeddings, consistently outperformed the MLP-based fusion by 1 percentage point in CRB and by 2 percentage points in DSMRB.

\subsection{Performance at 5-day Horizon}
\begin{figure}[htbp] 
  \centering
  \includegraphics[width=\columnwidth]{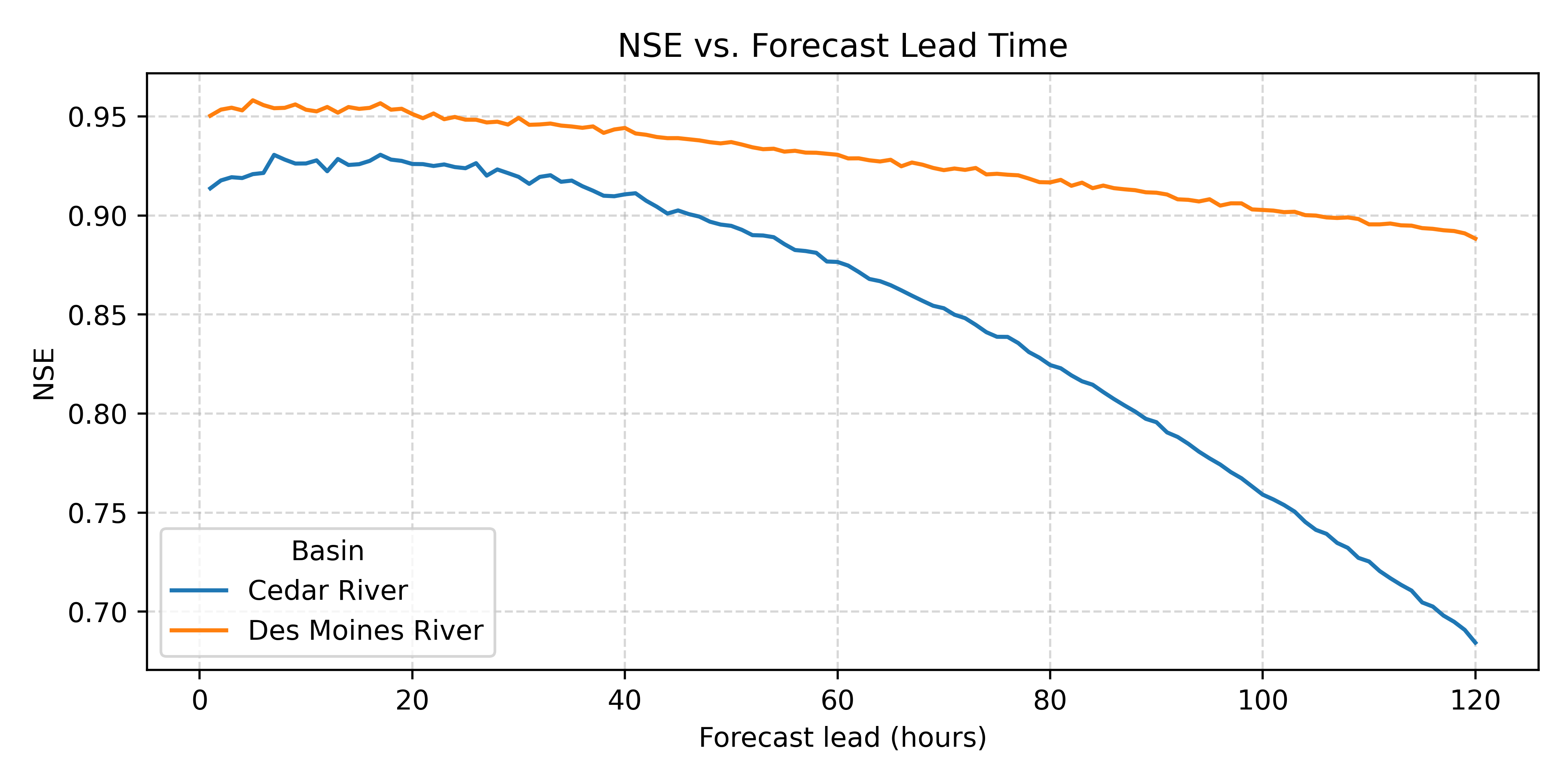}
  \caption{Basin-NSE over a 120-hour horizon using 72 hours of historical input. [\emph{Higher} is better]}\vspace{-1em}
  \label{fig:long-horizon}
\end{figure}
\noindent Fig.~\ref{fig:long-horizon} shows model performance over 120 hours of lead time using 72 hours of input. In DSMRB, NSE declines gradually from ~0.95 to ~0.89, indicating stable long-term accuracy. In contrast, CRB peaks near 0.93 at 12h but drops sharply to <0.70 by hour 120. The observed disparity in performance degradation between DSMRB and CRB can likely be attributed to the difference in hydrological regime and basin scale. DSMRB, being larger and more humid, exhibits smoother and more sustained flow responses that are easier to model over long lead times. In contrast, in CRB, prediction errors accumulate faster as the lead time grows, resulting in a sharper decline in NSE.

\subsection{Interpretability}
\vspace{-1.2em}
\subsubsection{\textbf{Spatial Attention}}\label{subsubsec:spatial-attention}
\begin{figure}[htbp]
  \centering
\includegraphics[width=\columnwidth,keepaspectratio]{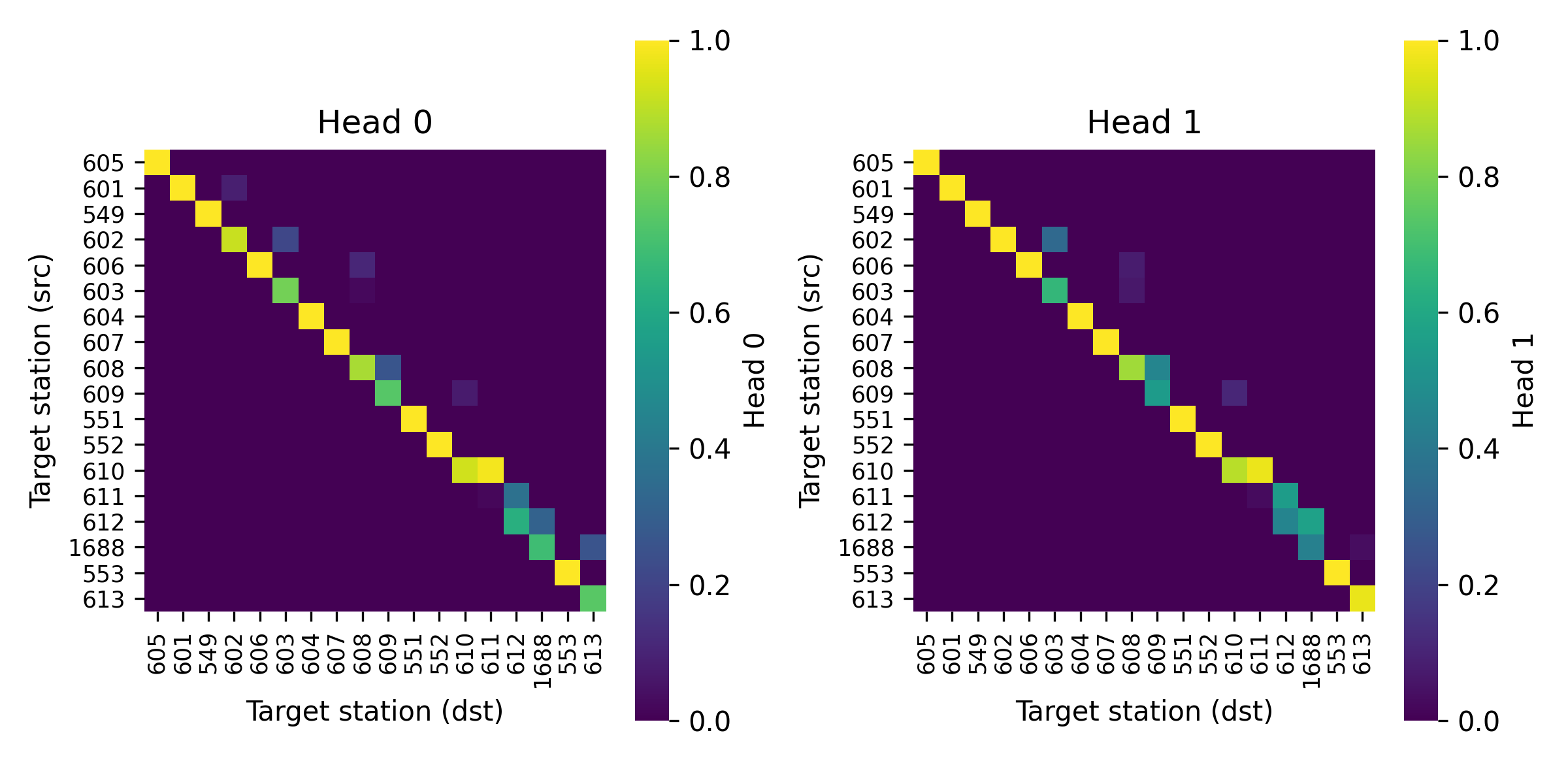}
  \caption{Model learned attention weights on catchment edges from source (y-axis) to destination (x-axis) stations across two heads. Bright yellow indicates stronger influence.}\vspace{-1em}
  \label{fig:attention_heads}
\end{figure}
We extracted attention maps from the trained model to investigate the learned spatial context. As shown in Fig. \ref{fig:attention_heads}, the model consistently focuses on the main upstream tributaries (see Fig. \ref{fig:cedar_river_full_graph}) while assigning near-zero weights to smaller side catchments. For instance, station 613 receives discharge from 553 and 1688, but the attention overwhelmingly favors 1688, effectively ignoring 553 after identifying it as an outlier. This selective weighting indicates that the GAT-GRU learns to emphasize the most informative hydrological routing during hidden state updates.

\subsubsection{\textbf{Temporal Attention}}\label{subsubsec:temporal-attention}
\begin{figure}[htbp]
  \centering
\includegraphics[width=\columnwidth,keepaspectratio]{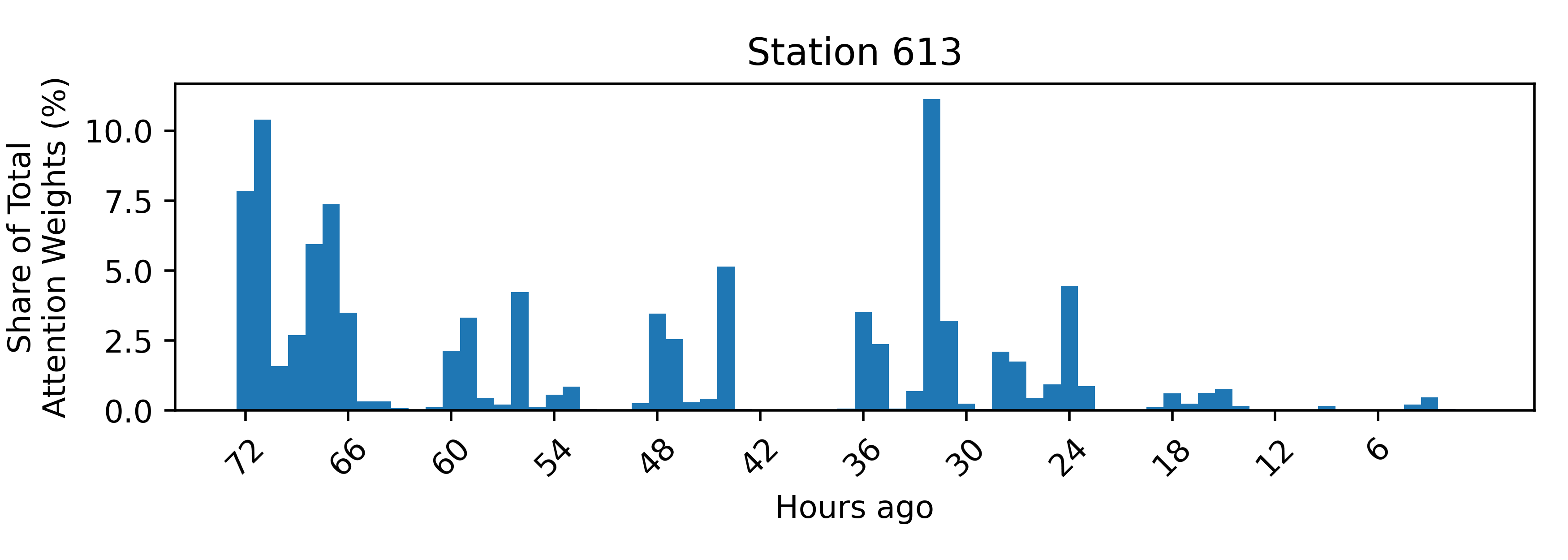}
  \caption{Distribution of temporal attention across past input timesteps in Station 613 in Cedar River Basin.}\vspace{-1em}
  \label{fig:temporal_attention}
\end{figure}
Similarly, Fig. \ref{fig:temporal_attention} shows the temporal attention distribution across the past 72-hour input window for station 613. The model attends not only to recent inputs (24-36 hours ago) but also to earlier periods ~66-72 hours and, to a lesser extent, to ~42-48 hours prior, indicating it has learned to focus on past inputs most hydrologically relevant for forecasting at the outlet.

\subsection{Scalability}
\begin{figure}[H]
  \centering
\includegraphics[width=0.9\columnwidth,keepaspectratio]{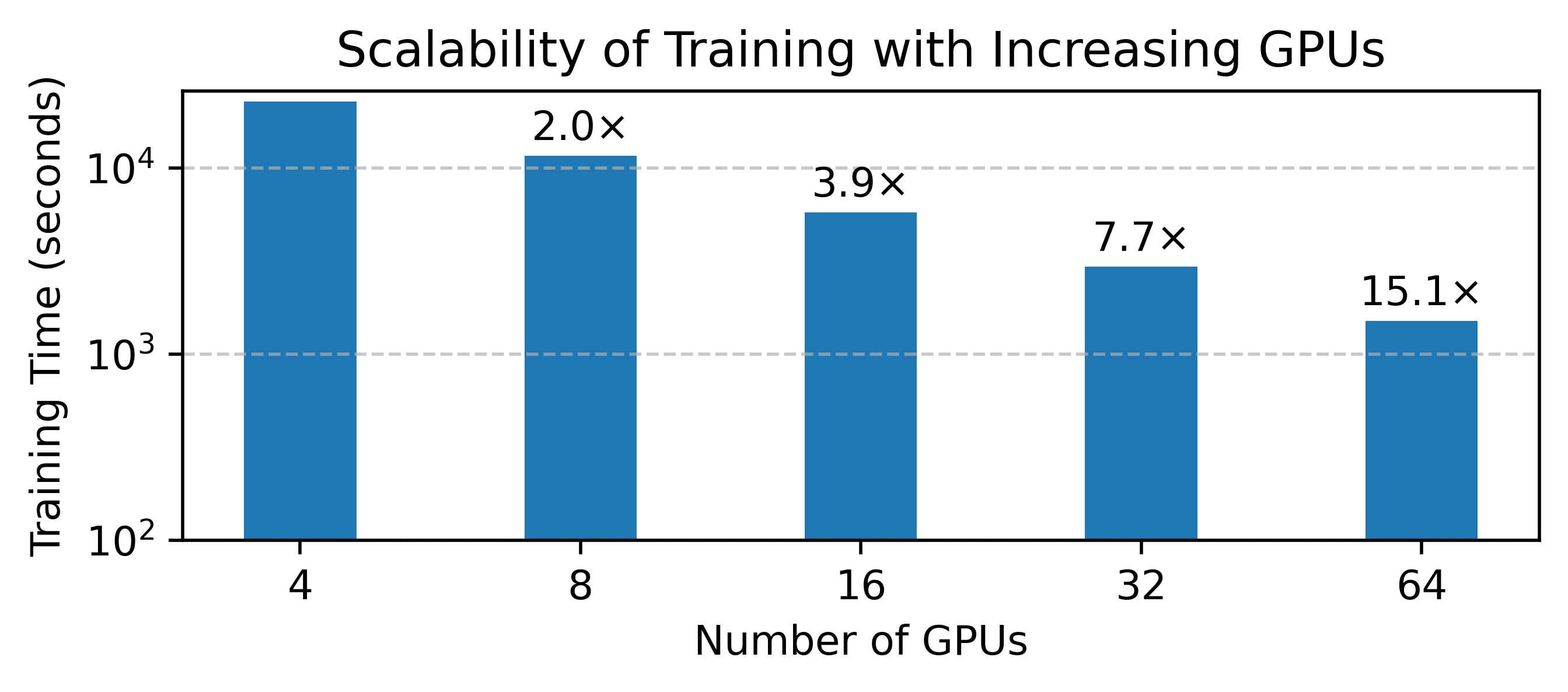}
  \caption{Scalability of distributed training in Cedar River Basin from 4-64 GPUs (1-16 machines, 4 GPUs/machine).}
  \label{fig:scalability}
\end{figure}

We evaluated the scalability of our model by measuring training time (models were run for 50 epochs) and increasing the \#GPUs from 4-64 (1-16 machines, 4 GPUs/machine) in CRB (Fig.~\ref{fig:scalability}). Training time decreases consistently, achieving 2-15$\times$ speedups compared to training on one machine. The reduction in training time is nearly linear up to 32 GPUs, and continues to improve at 64 GPUs, but with diminishing returns due to gradient communication overhead. These results validate the efficiency of our distributed pipeline in leveraging a multi-machine, multi-GPU distributed pipeline for large-scale spatiotemporal modeling.

\section{Conclusion}\label{sec:conclusion}
We presented HydroGAT, an interpretable, attention-based spatiotemporal model, combining spatial GAT and temporal transformer attention, that treats each river and land pixel as a node in a heterogeneous graph capturing both flow and inter-catchment dependencies. On two Midwestern U.S. basins, HydroGAT outperforms five baselines, with peak NSE 0.97 and KGE 0.96, in hourly discharge prediction. Ablation studies confirm the importance of each component in capturing runoff-routing dynamics. Our distributed, mixed-precision training pipeline scales efficiently to 64 GPUs with up to 15$\times$ speedup. The extracted attention maps offer interpretable insights into the spatial and temporal drivers in basins, making high-resolution, basin-wide flood forecasting feasible for real-world deployment. Additional environmental features (e.g., impervious surface, land use, vegetation) can be integrated to adapt to more heterogeneous basins. Beyond flood forecasting, HydroGAT can be readily applied to other spatiotemporal prediction tasks such as traffic flow or weather forecast, where both network topology and long-range temporal dependencies are critical.

\section{Acknowledgement}
This research was supported by the National Science Foundation (NSF) under Award 2243775 and utilized resources of the National Energy Research Scientific Computing Center (NERSC), a U.S. Department of Energy Office of Science User Facility, under NERSC award DOE-ERCAP 0029703.

\bibliographystyle{ACM-Reference-Format}
\bibliography{references}


\begin{thebibliography}{51}


\ifx \showCODEN    \undefined \def \showCODEN     #1{\unskip}     \fi
\ifx \showISBNx    \undefined \def \showISBNx     #1{\unskip}     \fi
\ifx \showISBNxiii \undefined \def \showISBNxiii  #1{\unskip}     \fi
\ifx \showISSN     \undefined \def \showISSN      #1{\unskip}     \fi
\ifx \showLCCN     \undefined \def \showLCCN      #1{\unskip}     \fi
\ifx \shownote     \undefined \def \shownote      #1{#1}          \fi
\ifx \showarticletitle \undefined \def \showarticletitle #1{#1}   \fi
\ifx \showURL      \undefined \def \showURL       {\relax}        \fi
\providecommand\bibfield[2]{#2}
\providecommand\bibinfo[2]{#2}
\providecommand\natexlab[1]{#1}
\providecommand\showeprint[2][]{arXiv:#2}

\bibitem[Alabbad et~al\mbox{.}(2023)]%
        {alabbad2023web}
\bibfield{author}{\bibinfo{person}{Yazeed Alabbad}, \bibinfo{person}{Enes Yildirim}, {and} \bibinfo{person}{Ibrahim Demir}.} \bibinfo{year}{2023}\natexlab{}.
\newblock \showarticletitle{A web-based analytical urban flood damage and loss estimation framework}.
\newblock \bibinfo{journal}{\emph{Environmental Modelling \& Software}}  \bibinfo{volume}{163} (\bibinfo{year}{2023}), \bibinfo{pages}{105670}.
\newblock


\bibitem[Bai and Tahmasebi(2023)]%
        {bai2023graph}
\bibfield{author}{\bibinfo{person}{Tao Bai} {and} \bibinfo{person}{Pejman Tahmasebi}.} \bibinfo{year}{2023}\natexlab{}.
\newblock \showarticletitle{Graph neural network for groundwater level forecasting}.
\newblock \bibinfo{journal}{\emph{Journal of Hydrology}}  \bibinfo{volume}{616} (\bibinfo{year}{2023}), \bibinfo{pages}{128792}.
\newblock


\bibitem[Bishop et~al\mbox{.}(1998)]%
        {bishop1998iowa}
\bibfield{author}{\bibinfo{person}{RA Bishop}, \bibinfo{person}{J Joens}, {and} \bibinfo{person}{J Zohrer}.} \bibinfo{year}{1998}\natexlab{}.
\newblock \showarticletitle{Iowa's wetlands, present and future with a focus on prairie potholes}.
\newblock \bibinfo{journal}{\emph{Journal of the Iowa Academy of Science: JIAS}} \bibinfo{volume}{105}, \bibinfo{number}{3} (\bibinfo{year}{1998}), \bibinfo{pages}{89--93}.
\newblock


\bibitem[Chung et~al\mbox{.}(2014)]%
        {chung2014empirical}
\bibfield{author}{\bibinfo{person}{Junyoung Chung}, \bibinfo{person}{Caglar Gulcehre}, \bibinfo{person}{KyungHyun Cho}, {and} \bibinfo{person}{Yoshua Bengio}.} \bibinfo{year}{2014}\natexlab{}.
\newblock \showarticletitle{Empirical evaluation of gated recurrent neural networks on sequence modeling}.
\newblock \bibinfo{journal}{\emph{arXiv preprint arXiv:1412.3555}} (\bibinfo{year}{2014}).
\newblock


\bibitem[Defferrard et~al\mbox{.}(2016)]%
        {defferrard2016convolutional}
\bibfield{author}{\bibinfo{person}{Micha{\"e}l Defferrard}, \bibinfo{person}{Xavier Bresson}, {and} \bibinfo{person}{Pierre Vandergheynst}.} \bibinfo{year}{2016}\natexlab{}.
\newblock \showarticletitle{Convolutional neural networks on graphs with fast localized spectral filtering}.
\newblock \bibinfo{journal}{\emph{Advances in neural information processing systems}}  \bibinfo{volume}{29} (\bibinfo{year}{2016}).
\newblock


\bibitem[Demir et~al\mbox{.}(2022)]%
        {demir2022waterbench}
\bibfield{author}{\bibinfo{person}{Ibrahim Demir}, \bibinfo{person}{Zhongrun Xiang}, \bibinfo{person}{Bekir Demiray}, {and} \bibinfo{person}{Muhammed Sit}.} \bibinfo{year}{2022}\natexlab{}.
\newblock \showarticletitle{Waterbench: a large-scale benchmark dataset for data-driven streamflow forecasting}.
\newblock \bibinfo{journal}{\emph{Earth System Science Data Discussions}}  \bibinfo{volume}{2022} (\bibinfo{year}{2022}), \bibinfo{pages}{1--19}.
\newblock


\bibitem[Du(2011)]%
        {du2011stageiv}
\bibfield{author}{\bibinfo{person}{Jianhua Du}.} \bibinfo{year}{2011}\natexlab{}.
\newblock \bibinfo{title}{NCEP/EMC 4KM Gridded Data (GRIB) Stage IV Data}.
\newblock \bibinfo{howpublished}{\url{https://doi.org/10.5065/D6PG1QDD}}.
\newblock


\bibitem[Fashae et~al\mbox{.}(2019)]%
        {fashae2019comparing}
\bibfield{author}{\bibinfo{person}{Olutoyin~A Fashae}, \bibinfo{person}{Adeyemi~O Olusola}, \bibinfo{person}{Ijeoma Ndubuisi}, {and} \bibinfo{person}{Christopher~Godwin Udomboso}.} \bibinfo{year}{2019}\natexlab{}.
\newblock \showarticletitle{Comparing ANN and ARIMA model in predicting the discharge of River Opeki from 2010 to 2020}.
\newblock \bibinfo{journal}{\emph{River research and applications}} \bibinfo{volume}{35}, \bibinfo{number}{2} (\bibinfo{year}{2019}), \bibinfo{pages}{169--177}.
\newblock


\bibitem[Feng et~al\mbox{.}(2022)]%
        {feng2022graph}
\bibfield{author}{\bibinfo{person}{Jun Feng}, \bibinfo{person}{Haichao Sha}, \bibinfo{person}{Yukai Ding}, \bibinfo{person}{Le Yan}, {and} \bibinfo{person}{Zhangheng Yu}.} \bibinfo{year}{2022}\natexlab{}.
\newblock \showarticletitle{Graph convolution based spatial-temporal attention LSTM model for flood forecasting}. In \bibinfo{booktitle}{\emph{2022 International Joint Conference on Neural Networks (IJCNN)}}. IEEE, \bibinfo{pages}{1--8}.
\newblock


\bibitem[Hochreiter(1998)]%
        {hochreiter1998vanishing}
\bibfield{author}{\bibinfo{person}{Sepp Hochreiter}.} \bibinfo{year}{1998}\natexlab{}.
\newblock \showarticletitle{The vanishing gradient problem during learning recurrent neural nets and problem solutions}.
\newblock \bibinfo{journal}{\emph{International Journal of Uncertainty, Fuzziness and Knowledge-Based Systems}} \bibinfo{volume}{6}, \bibinfo{number}{02} (\bibinfo{year}{1998}), \bibinfo{pages}{107--116}.
\newblock


\bibitem[Jia et~al\mbox{.}(2021)]%
        {jia2021physics}
\bibfield{author}{\bibinfo{person}{Xiaowei Jia}, \bibinfo{person}{Jacob Zwart}, \bibinfo{person}{Jeffrey Sadler}, \bibinfo{person}{Alison Appling}, \bibinfo{person}{Samantha Oliver}, \bibinfo{person}{Steven Markstrom}, \bibinfo{person}{Jared Willard}, \bibinfo{person}{Shaoming Xu}, \bibinfo{person}{Michael Steinbach}, \bibinfo{person}{Jordan Read}, {et~al\mbox{.}}} \bibinfo{year}{2021}\natexlab{}.
\newblock \showarticletitle{Physics-guided recurrent graph model for predicting flow and temperature in river networks}. In \bibinfo{booktitle}{\emph{Proceedings of the 2021 SIAM International Conference on Data Mining (SDM)}}. SIAM, \bibinfo{pages}{612--620}.
\newblock


\bibitem[Jiang et~al\mbox{.}(2022)]%
        {articleJiang}
\bibfield{author}{\bibinfo{person}{Jiange Jiang}, \bibinfo{person}{Zhan Liao}, \bibinfo{person}{Yang Zhou}, \bibinfo{person}{Hao Wang}, {and} \bibinfo{person}{Qingqi Pei}.} \bibinfo{year}{2022}\natexlab{}.
\newblock \showarticletitle{A short-term flood prediction based on spatial deep learning network: A case study for Xi County, China}.
\newblock \bibinfo{journal}{\emph{Journal of Hydrology}}  \bibinfo{volume}{607} (\bibinfo{date}{04} \bibinfo{year}{2022}), \bibinfo{pages}{127535}.
\newblock
\href{https://doi.org/10.1016/j.jhydrol.2022.127535}{doi:\nolinkurl{10.1016/j.jhydrol.2022.127535}}


\bibitem[Kipf and Welling(2016)]%
        {kipf2016semi}
\bibfield{author}{\bibinfo{person}{Thomas~N Kipf} {and} \bibinfo{person}{Max Welling}.} \bibinfo{year}{2016}\natexlab{}.
\newblock \showarticletitle{Semi-supervised classification with graph convolutional networks}.
\newblock \bibinfo{journal}{\emph{arXiv preprint arXiv:1609.02907}} (\bibinfo{year}{2016}).
\newblock


\bibitem[Kratzert et~al\mbox{.}(2021)]%
        {kratzert2021large}
\bibfield{author}{\bibinfo{person}{Frederik Kratzert}, \bibinfo{person}{Daniel Klotz}, \bibinfo{person}{Martin Gauch}, \bibinfo{person}{Christoph Klingler}, \bibinfo{person}{Grey Nearing}, {and} \bibinfo{person}{Sepp Hochreiter}.} \bibinfo{year}{2021}\natexlab{}.
\newblock \showarticletitle{Large-scale river network modeling using graph neural networks}. In \bibinfo{booktitle}{\emph{EGU General Assembly Conference Abstracts}}. \bibinfo{pages}{EGU21--13375}.
\newblock


\bibitem[Kumar et~al\mbox{.}(2024)]%
        {kumar2024data}
\bibfield{author}{\bibinfo{person}{Vikram Kumar}, \bibinfo{person}{Selim Unal}, \bibinfo{person}{Suraj~Kumar Bhagat}, {and} \bibinfo{person}{Tiyasha Tiyasha}.} \bibinfo{year}{2024}\natexlab{}.
\newblock \showarticletitle{A data-driven approach to river discharge forecasting in the Himalayan region: Insights from Aglar and Paligaad rivers}.
\newblock \bibinfo{journal}{\emph{Results in Engineering}}  \bibinfo{volume}{22} (\bibinfo{year}{2024}), \bibinfo{pages}{102044}.
\newblock


\bibitem[Li et~al\mbox{.}(2018)]%
        {li2018diffusion}
\bibfield{author}{\bibinfo{person}{Yaguang Li}, \bibinfo{person}{Rose Yu}, \bibinfo{person}{Cyrus Shahabi}, {and} \bibinfo{person}{Yan Liu}.} \bibinfo{year}{2018}\natexlab{}.
\newblock \showarticletitle{Diffusion Convolutional Recurrent Neural Network: Data-Driven Traffic Forecasting}. In \bibinfo{booktitle}{\emph{International Conference on Learning Representations (ICLR)}}.
\newblock


\bibitem[Loshchilov and Hutter(2017)]%
        {loshchilov2017decoupled}
\bibfield{author}{\bibinfo{person}{Ilya Loshchilov} {and} \bibinfo{person}{Frank Hutter}.} \bibinfo{year}{2017}\natexlab{}.
\newblock \showarticletitle{Decoupled weight decay regularization}.
\newblock \bibinfo{journal}{\emph{arXiv preprint arXiv:1711.05101}} (\bibinfo{year}{2017}).
\newblock


\bibitem[Mialon et~al\mbox{.}(2021)]%
        {mialon2021graphit}
\bibfield{author}{\bibinfo{person}{Gr{\'e}goire Mialon}, \bibinfo{person}{Dexiong Chen}, \bibinfo{person}{Margot Selosse}, {and} \bibinfo{person}{Julien Mairal}.} \bibinfo{year}{2021}\natexlab{}.
\newblock \showarticletitle{Graphit: Encoding graph structure in transformers}.
\newblock \bibinfo{journal}{\emph{arXiv preprint arXiv:2106.05667}} (\bibinfo{year}{2021}).
\newblock


\bibitem[Mostafa(2021)]%
        {mostafa2021sequential}
\bibfield{author}{\bibinfo{person}{Hesham Mostafa}.} \bibinfo{year}{2021}\natexlab{}.
\newblock \showarticletitle{Sequential Aggregation and Rematerialization: Distributed Full-batch Training of Graph Neural Networks on Large Graphs. arXiv 2021}.
\newblock \bibinfo{journal}{\emph{arXiv preprint arXiv:2111.06483}} (\bibinfo{year}{2021}).
\newblock


\bibitem[Otchere et~al\mbox{.}(2021)]%
        {otchere2021application}
\bibfield{author}{\bibinfo{person}{Daniel~Asante Otchere}, \bibinfo{person}{Tarek Omar~Arbi Ganat}, \bibinfo{person}{Raoof Gholami}, {and} \bibinfo{person}{Syahrir Ridha}.} \bibinfo{year}{2021}\natexlab{}.
\newblock \showarticletitle{Application of supervised machine learning paradigms in the prediction of petroleum reservoir properties: Comparative analysis of ANN and SVM models}.
\newblock \bibinfo{journal}{\emph{Journal of Petroleum Science and Engineering}}  \bibinfo{volume}{200} (\bibinfo{year}{2021}), \bibinfo{pages}{108182}.
\newblock


\bibitem[Ruiz et~al\mbox{.}(2019)]%
        {ruiz2019gatedgraphconvolutionalrecurrent}
\bibfield{author}{\bibinfo{person}{Luana Ruiz}, \bibinfo{person}{Fernando Gama}, {and} \bibinfo{person}{Alejandro Ribeiro}.} \bibinfo{year}{2019}\natexlab{}.
\newblock \bibinfo{title}{Gated Graph Convolutional Recurrent Neural Networks}.
\newblock
\showeprint[arxiv]{1903.01888}~[cs.LG]
\urldef\tempurl%
\url{https://arxiv.org/abs/1903.01888}
\showURL{%
\tempurl}


\bibitem[Sarkar et~al\mbox{.}(2023)]%
        {sarkar2023}
\bibfield{author}{\bibinfo{person}{Aishwarya Sarkar}, \bibinfo{person}{Jien Zhang}, \bibinfo{person}{Chaoqun Lu}, {and} \bibinfo{person}{Ali Jannesari}.} \bibinfo{year}{2023}\natexlab{}.
\newblock \showarticletitle{Domain-Aware Scalable Distributed Training for Geo-Spatiotemporal Data}. In \bibinfo{booktitle}{\emph{2023 IEEE International Parallel and Distributed Processing Symposium Workshops (IPDPSW)}}. \bibinfo{pages}{960--969}.
\newblock
\href{https://doi.org/10.1109/IPDPSW59300.2023.00159}{doi:\nolinkurl{10.1109/IPDPSW59300.2023.00159}}


\bibitem[Sit et~al\mbox{.}(2021)]%
        {sit2021shorttermhourlystreamflowprediction}
\bibfield{author}{\bibinfo{person}{Muhammed Sit}, \bibinfo{person}{Bekir Demiray}, {and} \bibinfo{person}{Ibrahim Demir}.} \bibinfo{year}{2021}\natexlab{}.
\newblock \bibinfo{title}{Short-term Hourly Streamflow Prediction with Graph Convolutional GRU Networks}.
\newblock
\showeprint[arxiv]{2107.07039}~[cs.LG]
\urldef\tempurl%
\url{https://arxiv.org/abs/2107.07039}
\showURL{%
\tempurl}


\bibitem[Sit et~al\mbox{.}(2023)]%
        {sit2023spatial}
\bibfield{author}{\bibinfo{person}{Muhammed Sit}, \bibinfo{person}{Bekir~Zahit Demiray}, {and} \bibinfo{person}{Ibrahim Demir}.} \bibinfo{year}{2023}\natexlab{}.
\newblock \showarticletitle{Spatial downscaling of streamflow data with attention based spatio-temporal graph convolutional networks}.
\newblock \bibinfo{journal}{\emph{EarthArxiv}} (\bibinfo{year}{2023}).
\newblock


\bibitem[Sit et~al\mbox{.}(2020)]%
        {sit2020comprehensive}
\bibfield{author}{\bibinfo{person}{Muhammed Sit}, \bibinfo{person}{Bekir~Z Demiray}, \bibinfo{person}{Zhongrun Xiang}, \bibinfo{person}{Gregory~J Ewing}, \bibinfo{person}{Yusuf Sermet}, {and} \bibinfo{person}{Ibrahim Demir}.} \bibinfo{year}{2020}\natexlab{}.
\newblock \showarticletitle{A comprehensive review of deep learning applications in hydrology and water resources}.
\newblock \bibinfo{journal}{\emph{Water Science and Technology}} \bibinfo{volume}{82}, \bibinfo{number}{12} (\bibinfo{year}{2020}), \bibinfo{pages}{2635--2670}.
\newblock


\bibitem[Stalder et~al\mbox{.}(2021)]%
        {stalder2021probabilistic}
\bibfield{author}{\bibinfo{person}{Michael Stalder}, \bibinfo{person}{Firat Ozdemir}, \bibinfo{person}{Artur Safin}, \bibinfo{person}{Jonas Sukys}, \bibinfo{person}{Damien Bouffard}, {and} \bibinfo{person}{Fernando Perez-Cruz}.} \bibinfo{year}{2021}\natexlab{}.
\newblock \showarticletitle{Probabilistic modeling of lake surface water temperature using a Bayesian spatio-temporal graph convolutional neural network}.
\newblock \bibinfo{journal}{\emph{arXiv preprint arXiv:2109.13235}} (\bibinfo{year}{2021}).
\newblock


\bibitem[Sun et~al\mbox{.}(2021)]%
        {sun2021explore}
\bibfield{author}{\bibinfo{person}{Alexander~Y Sun}, \bibinfo{person}{Peishi Jiang}, \bibinfo{person}{Maruti~K Mudunuru}, {and} \bibinfo{person}{Xingyuan Chen}.} \bibinfo{year}{2021}\natexlab{}.
\newblock \showarticletitle{Explore spatio-temporal learning of large sample hydrology using graph neural networks}.
\newblock \bibinfo{journal}{\emph{Water Resources Research}} \bibinfo{volume}{57}, \bibinfo{number}{12} (\bibinfo{year}{2021}), \bibinfo{pages}{e2021WR030394}.
\newblock


\bibitem[Sun et~al\mbox{.}(2022)]%
        {sun2022graph}
\bibfield{author}{\bibinfo{person}{Alexander~Y Sun}, \bibinfo{person}{Peishi Jiang}, \bibinfo{person}{Zong-Liang Yang}, \bibinfo{person}{Yangxinyu Xie}, {and} \bibinfo{person}{Xingyuan Chen}.} \bibinfo{year}{2022}\natexlab{}.
\newblock \showarticletitle{A graph neural network approach to basin-scale river network learning: The role of physics-based connectivity and data fusion}.
\newblock \bibinfo{journal}{\emph{Hydrology and Earth System Sciences Discussions}}  \bibinfo{volume}{2022} (\bibinfo{year}{2022}), \bibinfo{pages}{1--35}.
\newblock


\bibitem[Survey(2021)]%
        {usgs2021dem}
\bibfield{author}{\bibinfo{person}{United States~Geological Survey}.} \bibinfo{year}{2021}\natexlab{}.
\newblock \bibinfo{title}{3D Elevation Program 1 arc-second Digital Elevation Model}.
\newblock
\href{https://doi.org/10.5069/G98K778D}{doi:\nolinkurl{10.5069/G98K778D}}


\bibitem[Valayamkunnath et~al\mbox{.}(2020)]%
        {valayamkunnath2020mapping}
\bibfield{author}{\bibinfo{person}{Prasanth Valayamkunnath}, \bibinfo{person}{Michael Barlage}, \bibinfo{person}{Fei Chen}, \bibinfo{person}{David~J Gochis}, {and} \bibinfo{person}{Kristie~J Franz}.} \bibinfo{year}{2020}\natexlab{}.
\newblock \showarticletitle{Mapping of 30-meter resolution tile-drained croplands using a geospatial modeling approach}.
\newblock \bibinfo{journal}{\emph{Scientific data}} \bibinfo{volume}{7}, \bibinfo{number}{1} (\bibinfo{year}{2020}), \bibinfo{pages}{257}.
\newblock


\bibitem[Vaswani et~al\mbox{.}(2017)]%
        {vaswani2017attention}
\bibfield{author}{\bibinfo{person}{Ashish Vaswani}, \bibinfo{person}{Noam Shazeer}, \bibinfo{person}{Niki Parmar}, \bibinfo{person}{Jakob Uszkoreit}, \bibinfo{person}{Llion Jones}, \bibinfo{person}{Aidan~N Gomez}, \bibinfo{person}{{\L}ukasz Kaiser}, {and} \bibinfo{person}{Illia Polosukhin}.} \bibinfo{year}{2017}\natexlab{}.
\newblock \showarticletitle{Attention is all you need}.
\newblock \bibinfo{journal}{\emph{Advances in neural information processing systems}}  \bibinfo{volume}{30} (\bibinfo{year}{2017}).
\newblock


\bibitem[Veli{\v{c}}kovi{\'c} et~al\mbox{.}(2017)]%
        {velivckovic2017graph}
\bibfield{author}{\bibinfo{person}{Petar Veli{\v{c}}kovi{\'c}}, \bibinfo{person}{Guillem Cucurull}, \bibinfo{person}{Arantxa Casanova}, \bibinfo{person}{Adriana Romero}, \bibinfo{person}{Pietro Lio}, {and} \bibinfo{person}{Yoshua Bengio}.} \bibinfo{year}{2017}\natexlab{}.
\newblock \showarticletitle{Graph attention networks}.
\newblock \bibinfo{journal}{\emph{arXiv preprint arXiv:1710.10903}} (\bibinfo{year}{2017}).
\newblock


\bibitem[Wang et~al\mbox{.}(2024)]%
        {article}
\bibfield{author}{\bibinfo{person}{Chao Wang}, \bibinfo{person}{Shijie Jiang}, \bibinfo{person}{yi Zheng}, \bibinfo{person}{Feng Han}, \bibinfo{person}{Rohini Kumar}, \bibinfo{person}{Oldrich Rakovec}, {and} \bibinfo{person}{Siqi Li}.} \bibinfo{year}{2024}\natexlab{}.
\newblock \showarticletitle{Distributed Hydrological Modeling With Physics‐Encoded Deep Learning: A General Framework and Its Application in the Amazon}.
\newblock \bibinfo{journal}{\emph{Water Resources Research}}  \bibinfo{volume}{60} (\bibinfo{date}{04} \bibinfo{year}{2024}).
\newblock
\href{https://doi.org/10.1029/2023WR036170}{doi:\nolinkurl{10.1029/2023WR036170}}


\bibitem[Waseem et~al\mbox{.}(2017)]%
        {waseem2017review}
\bibfield{author}{\bibinfo{person}{M Waseem}, \bibinfo{person}{N Mani}, \bibinfo{person}{G Andiego}, {and} \bibinfo{person}{M Usman}.} \bibinfo{year}{2017}\natexlab{}.
\newblock \showarticletitle{A review of criteria of fit for hydrological models}.
\newblock \bibinfo{journal}{\emph{International Research Journal of Engineering and Technology (IRJET)}} \bibinfo{volume}{4}, \bibinfo{number}{11} (\bibinfo{year}{2017}), \bibinfo{pages}{1765--1772}.
\newblock


\bibitem[Willard et~al\mbox{.}(2025)]%
        {willard2025time}
\bibfield{author}{\bibinfo{person}{Jared~D Willard}, \bibinfo{person}{Charuleka Varadharajan}, \bibinfo{person}{Xiaowei Jia}, {and} \bibinfo{person}{Vipin Kumar}.} \bibinfo{year}{2025}\natexlab{}.
\newblock \showarticletitle{Time series predictions in unmonitored sites: A survey of machine learning techniques in water resources}.
\newblock \bibinfo{journal}{\emph{Environmental Data Science}}  \bibinfo{volume}{4} (\bibinfo{year}{2025}), \bibinfo{pages}{e7}.
\newblock


\bibitem[Wong et~al\mbox{.}(2020)]%
        {wong2020flood}
\bibfield{author}{\bibinfo{person}{Wei~Ming Wong}, \bibinfo{person}{Siva~Kumar Subramaniam}, \bibinfo{person}{Farah~Shahnaz Feroz}, \bibinfo{person}{Indra~Devi Subramaniam}, {and} \bibinfo{person}{Lew Ai~Fen Rose}.} \bibinfo{year}{2020}\natexlab{}.
\newblock \showarticletitle{Flood prediction using ARIMA model in Sungai Melaka, Malaysia}.
\newblock \bibinfo{journal}{\emph{International Journal}} \bibinfo{volume}{9}, \bibinfo{number}{4} (\bibinfo{year}{2020}).
\newblock


\bibitem[Wu et~al\mbox{.}(2021)]%
        {wu2021representing}
\bibfield{author}{\bibinfo{person}{Zhanghao Wu}, \bibinfo{person}{Paras Jain}, \bibinfo{person}{Matthew Wright}, \bibinfo{person}{Azalia Mirhoseini}, \bibinfo{person}{Joseph~E Gonzalez}, {and} \bibinfo{person}{Ion Stoica}.} \bibinfo{year}{2021}\natexlab{}.
\newblock \showarticletitle{Representing long-range context for graph neural networks with global attention}.
\newblock \bibinfo{journal}{\emph{Advances in neural information processing systems}}  \bibinfo{volume}{34} (\bibinfo{year}{2021}), \bibinfo{pages}{13266--13279}.
\newblock


\bibitem[Wu et~al\mbox{.}(2019)]%
        {wu2019graph}
\bibfield{author}{\bibinfo{person}{Zhijian Wu}, \bibinfo{person}{Shirui Pan}, \bibinfo{person}{Fengwen Chen}, \bibinfo{person}{Guodong Long}, \bibinfo{person}{Chengqi Zhang}, {and} \bibinfo{person}{Philip~S Yu}.} \bibinfo{year}{2019}\natexlab{}.
\newblock \showarticletitle{Graph WaveNet for Deep Spatial-Temporal Graph Modeling}. In \bibinfo{booktitle}{\emph{IJCAI}}.
\newblock


\bibitem[Xiang and Demir(2021)]%
        {xiang2021high}
\bibfield{author}{\bibinfo{person}{Zhongrun Xiang} {and} \bibinfo{person}{Ibrahim Demir}.} \bibinfo{year}{2021}\natexlab{}.
\newblock \showarticletitle{High-resolution rainfall-runoff modeling using graph neural network}.
\newblock \bibinfo{journal}{\emph{arXiv preprint arXiv:2110.10833}} (\bibinfo{year}{2021}).
\newblock


\bibitem[Xiang and Demir(2022)]%
        {xiang2022fully}
\bibfield{author}{\bibinfo{person}{Zhongrun Xiang} {and} \bibinfo{person}{Ibrahim Demir}.} \bibinfo{year}{2022}\natexlab{}.
\newblock \showarticletitle{Fully distributed rainfall-runoff modeling using spatial-temporal graph neural network}.
\newblock \bibinfo{journal}{\emph{EarthArxiv}} (\bibinfo{year}{2022}).
\newblock


\bibitem[Yang and Deslippe(2020)]%
        {yang2020accelerate}
\bibfield{author}{\bibinfo{person}{Charlene Yang} {and} \bibinfo{person}{Jack Deslippe}.} \bibinfo{year}{2020}\natexlab{}.
\newblock \showarticletitle{Accelerate science on perlmutter with nersc}.
\newblock \bibinfo{journal}{\emph{Bulletin of the American Physical Society}}  \bibinfo{volume}{65} (\bibinfo{year}{2020}).
\newblock


\bibitem[Yang et~al\mbox{.}(2023a)]%
        {yang2023runoff}
\bibfield{author}{\bibinfo{person}{Shuai Yang}, \bibinfo{person}{Yueqin Zhang}, {and} \bibinfo{person}{Zehua Zhang}.} \bibinfo{year}{2023}\natexlab{a}.
\newblock \showarticletitle{Runoff prediction based on dynamic spatiotemporal graph neural network}.
\newblock \bibinfo{journal}{\emph{Water}} \bibinfo{volume}{15}, \bibinfo{number}{13} (\bibinfo{year}{2023}), \bibinfo{pages}{2463}.
\newblock


\bibitem[Yang et~al\mbox{.}(2023b)]%
        {yang2023study}
\bibfield{author}{\bibinfo{person}{Weichao Yang}, \bibinfo{person}{Chuanxing Zheng}, \bibinfo{person}{Xuelian Jiang}, \bibinfo{person}{Hao Wang}, \bibinfo{person}{Jijian Lian}, \bibinfo{person}{De Hu}, {and} \bibinfo{person}{Airong Zheng}.} \bibinfo{year}{2023}\natexlab{b}.
\newblock \showarticletitle{Study on urban flood simulation based on a novel model of SWTM coupling D8 flow direction and backflow effect}.
\newblock \bibinfo{journal}{\emph{Journal of Hydrology}}  \bibinfo{volume}{621} (\bibinfo{year}{2023}), \bibinfo{pages}{129608}.
\newblock


\bibitem[Yildirim et~al\mbox{.}(2022)]%
        {yildirim2022flood}
\bibfield{author}{\bibinfo{person}{Enes Yildirim}, \bibinfo{person}{Craig Just}, {and} \bibinfo{person}{Ibrahim Demir}.} \bibinfo{year}{2022}\natexlab{}.
\newblock \showarticletitle{Flood risk assessment and quantification at the community and property level in the State of Iowa}.
\newblock \bibinfo{journal}{\emph{International journal of disaster risk reduction}}  \bibinfo{volume}{77} (\bibinfo{year}{2022}), \bibinfo{pages}{103106}.
\newblock


\bibitem[Ying et~al\mbox{.}(2021)]%
        {ying2021transformers}
\bibfield{author}{\bibinfo{person}{Chengxuan Ying}, \bibinfo{person}{Tianle Cai}, \bibinfo{person}{Shengjie Luo}, \bibinfo{person}{Shuxin Zheng}, \bibinfo{person}{Guolin Ke}, \bibinfo{person}{Di He}, \bibinfo{person}{Yanming Shen}, {and} \bibinfo{person}{Tie-Yan Liu}.} \bibinfo{year}{2021}\natexlab{}.
\newblock \showarticletitle{Do transformers really perform badly for graph representation?}
\newblock \bibinfo{journal}{\emph{Advances in neural information processing systems}}  \bibinfo{volume}{34} (\bibinfo{year}{2021}), \bibinfo{pages}{28877--28888}.
\newblock


\bibitem[Yu et~al\mbox{.}(2017)]%
        {yu2017spatio}
\bibfield{author}{\bibinfo{person}{Bing Yu}, \bibinfo{person}{Haoteng Yin}, {and} \bibinfo{person}{Zhanxing Zhu}.} \bibinfo{year}{2017}\natexlab{}.
\newblock \showarticletitle{Spatio-temporal graph convolutional networks: A deep learning framework for traffic forecasting}.
\newblock \bibinfo{journal}{\emph{arXiv preprint arXiv:1709.04875}} (\bibinfo{year}{2017}).
\newblock


\bibitem[Yu et~al\mbox{.}(2020)]%
        {yu2020spatio}
\bibfield{author}{\bibinfo{person}{Cunjun Yu}, \bibinfo{person}{Xiao Ma}, \bibinfo{person}{Jiawei Ren}, \bibinfo{person}{Haiyu Zhao}, {and} \bibinfo{person}{Shuai Yi}.} \bibinfo{year}{2020}\natexlab{}.
\newblock \showarticletitle{Spatio-temporal graph transformer networks for pedestrian trajectory prediction}. In \bibinfo{booktitle}{\emph{European conference on computer vision}}. Springer, \bibinfo{pages}{507--523}.
\newblock


\bibitem[Yu et~al\mbox{.}(2022)]%
        {yu2022recognition}
\bibfield{author}{\bibinfo{person}{Huafei Yu}, \bibinfo{person}{Tinghua Ai}, \bibinfo{person}{Min Yang}, \bibinfo{person}{Lina Huang}, {and} \bibinfo{person}{Jiaming Yuan}.} \bibinfo{year}{2022}\natexlab{}.
\newblock \showarticletitle{A recognition method for drainage patterns using a graph convolutional network}.
\newblock \bibinfo{journal}{\emph{International Journal of Applied Earth Observation and Geoinformation}}  \bibinfo{volume}{107} (\bibinfo{year}{2022}), \bibinfo{pages}{102696}.
\newblock


\bibitem[Yuan et~al\mbox{.}(2023)]%
        {yuan2023comprehensive}
\bibfield{author}{\bibinfo{person}{Hao Yuan}, \bibinfo{person}{Yajiong Liu}, \bibinfo{person}{Yanfeng Zhang}, \bibinfo{person}{Xin Ai}, \bibinfo{person}{Qiange Wang}, \bibinfo{person}{Chaoyi Chen}, \bibinfo{person}{Yu Gu}, {and} \bibinfo{person}{Ge Yu}.} \bibinfo{year}{2023}\natexlab{}.
\newblock \showarticletitle{Comprehensive evaluation of gnn training systems: A data management perspective}.
\newblock \bibinfo{journal}{\emph{arXiv preprint arXiv:2311.13279}} (\bibinfo{year}{2023}).
\newblock


\bibitem[Zanfei et~al\mbox{.}(2022)]%
        {zanfei2022graph}
\bibfield{author}{\bibinfo{person}{Ariele Zanfei}, \bibinfo{person}{Bruno~M Brentan}, \bibinfo{person}{Andrea Menapace}, \bibinfo{person}{Maurizio Righetti}, {and} \bibinfo{person}{Manuel Herrera}.} \bibinfo{year}{2022}\natexlab{}.
\newblock \showarticletitle{Graph convolutional recurrent neural networks for water demand forecasting}.
\newblock \bibinfo{journal}{\emph{Water Resources Research}} \bibinfo{volume}{58}, \bibinfo{number}{7} (\bibinfo{year}{2022}), \bibinfo{pages}{e2022WR032299}.
\newblock


\bibitem[Zhao et~al\mbox{.}(2020)]%
        {zhao2020joint}
\bibfield{author}{\bibinfo{person}{Qun Zhao}, \bibinfo{person}{Yuelong Zhu}, \bibinfo{person}{Kai Shu}, \bibinfo{person}{Dingsheng Wan}, \bibinfo{person}{Yufeng Yu}, \bibinfo{person}{Xudong Zhou}, {and} \bibinfo{person}{Huan Liu}.} \bibinfo{year}{2020}\natexlab{}.
\newblock \showarticletitle{Joint spatial and temporal modeling for hydrological prediction}.
\newblock \bibinfo{journal}{\emph{Ieee Access}}  \bibinfo{volume}{8} (\bibinfo{year}{2020}), \bibinfo{pages}{78492--78503}.
\newblock


\end{thebibliography}

\end{document}